\pdfoutput=1

\documentclass[11pt]{article}

\usepackage[final]{acl}

\usepackage{times}
\usepackage{latexsym}

\usepackage[T1]{fontenc}

\usepackage[utf8]{inputenc}

\usepackage{microtype}

\usepackage{inconsolata}

\usepackage{graphicx}

\usepackage{amsmath,amsfonts,bm}

\def\eqref#1{equation~\ref{#1}}

\def\1{\bm{1}}

\DeclareMathAlphabet{\mathsfit}{\encodingdefault}{\sfdefault}{m}{sl}
\SetMathAlphabet{\mathsfit}{bold}{\encodingdefault}{\sfdefault}{bx}{n}

\DeclareMathOperator*{\argmin}{arg\,min}

\usepackage{enumitem}
\usepackage{wrapfig}
\usepackage[utf8]{inputenc} %
\usepackage[T1]{fontenc}    %
\usepackage{hyperref}       %
\usepackage{url}            %
\usepackage{booktabs}       %
\usepackage{amsfonts}       %
\usepackage{nicefrac}       %
\usepackage{xcolor}         %
\usepackage{caption}
\usepackage{amsmath}
\usepackage{amssymb}
\usepackage{amsthm}
\usepackage{subcaption}
\usepackage{devanagari}
\usepackage{multirow}
\usepackage{arydshln}
\usepackage{tablefootnote}
\usepackage{pifont}%

\newcommand{\circled}[1]{\raisebox{.5pt}{\textcircled{\raisebox{-.9pt} {#1}}}}

\usepackage[disable]{todonotes}

\makeatletter
\def\adl@drawiv#1#2#3{%
        \hskip.5\tabcolsep
        \xleaders#3{#2.5\@tempdimb #1{1}#2.5\@tempdimb}%
                #2\z@ plus1fil minus1fil\relax
        \hskip.5\tabcolsep}
\newcommand{\cdashlinelr}[1]{%
  \noalign{\vskip\aboverulesep
           \global\let\@dashdrawstore\adl@draw
           \global\let\adl@draw\adl@drawiv}
  \cdashline{#1}
  \noalign{\global\let\adl@draw\@dashdrawstore
           \vskip\belowrulesep}}
\makeatother

\title{Textless Speech-to-Speech Translation With Limited Parallel Data}

\author{Anuj Diwan$^\diamondsuit$, Anirudh Srinivasan$^\diamondsuit$, David Harwath$^\diamondsuit$, Eunsol Choi$^\clubsuit$ \\ $^\diamondsuit$ Department of Computer Science, The University of Texas at Austin \\
$^\clubsuit$ Department of Computer Science, New York University
\\ \texttt{\{anuj.diwan, anirudhs, harwath\}@utexas.edu}\hfill \texttt{eunsol@nyu.edu}}

\begin{document}
\maketitle
\begin{abstract}
Existing speech-to-speech translation (S2ST) models fall into two camps: they either leverage text as an intermediate step or require hundreds of hours of parallel speech data. Both approaches are incompatible with textless languages or language pairs with limited parallel data. We present \texttt{PFB}, a framework for training textless S2ST models that require just dozens of hours of parallel speech data. We first pretrain a model on large-scale monolingual speech data, finetune it with a small amount of parallel speech data ($20-60$ hours), and lastly train with an unsupervised backtranslation objective. We train and evaluate our models for English-to-German, German-to-English and Marathi-to-English translation on three different domains (European Parliament, Common Voice, and All India Radio) with single-speaker synthesized speech. Evaluated using the ASR-BLEU metric, our models achieve reasonable performance on all three domains, with some being within $1-2$ points of our higher-resourced topline.
\end{abstract}

\section{Introduction}
\label{sec:intro}
Speech-to-speech translation (S2ST) systems map input speech in the source language to output speech in the target language. In many ways, S2ST represents the \textit{holy grail} of translation as it enables natural, real-time, spoken communication. S2ST has a rich history, from cascaded systems combining Automatic Speech Recognition (ASR), Machine Translation (MT), and Text To Speech (TTS) technologies~\citep{1597243} to recently proposed neural end-to-end systems~\citep{lee-etal-2022-direct, seamlessm4t2023} that directly map from input source language speech to output target language speech. These systems~\citep{Jia2019DirectST,lee-etal-2022-direct,lee-etal-2022-textless,translatotron2, duquenne2022speechmatrix, seamlessm4t2023} have benefited from model and data scaling, leveraging increasing amounts of parallel speech and/or text data across languages. Yet, this is feasible only for a fraction of the world's 7000 languages~\citep{ethnologue}; the majority of world languages have low-resource or no parallel translation data available~\citep{haddow-etal-2022-survey}. Furthermore, thousands of languages are primarily spoken without standardized writing systems (about 3000 languages in Ethnologue~\citep{ethnologue} have no reported writing system), necessitating textless language processing.

Recent work on textless speech translation~\citep{lee-etal-2022-textless,kim2023manytomany} train end-to-end models on large amounts of parallel speech data, which is expensive to collect and makes these approaches difficult to adapt for low-resource speech translation. Sentence embedding-based modular approaches~\citep{Duquenne2022TModulesTM,duquenne2023sonarsentencelevelmultimodallanguageagnostic} do not require any parallel speech data, but still require parallel text data to learn cross-lingual sentence embedding spaces. On the other hand, unsupervised S2ST approaches~\citep{https://doi.org/10.48550/arxiv.2210.10191, fu2023improving, nachmani2023translatotron} do not need any parallel speech or text data at all, instead relying on unsupervised cross-lingual learning using large amounts of monolingual speech and text datasets. However, they either train cascaded models that have intermediate text outputs or end-to-end models that use text supervision during training. As a result, they are difficult to adapt for textless languages that are spoken, have non-standard orthographies or poor ASR systems.

\begin{figure*}
    \centering
    \includegraphics[width=\textwidth]{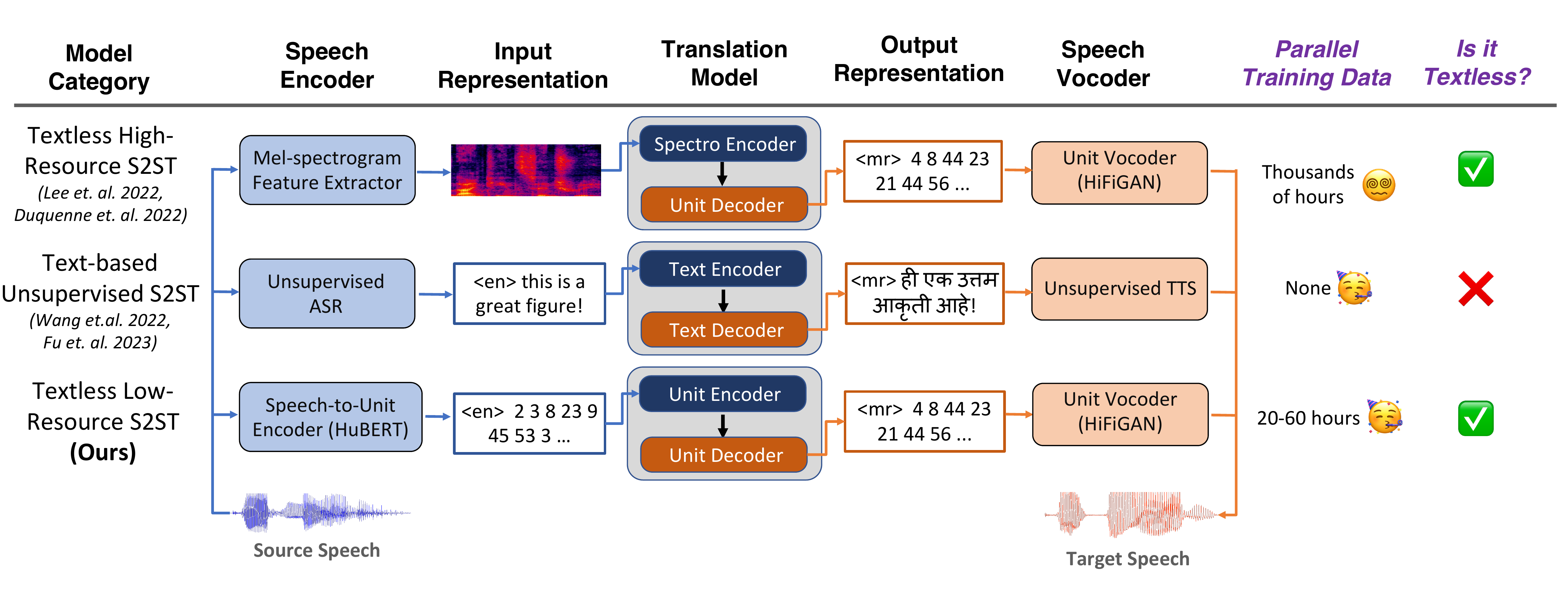}
    \caption{Overview of speech-to-speech translation systems. We compare our formulation to two relevant lines of work. We present the first textless speech-to-speech system that does not require a large parallel training dataset.}
    \label{fig:intro}
\end{figure*}

In this work, we adapt the unsupervised S2ST pipeline to work in a fully textless manner for the first time. We formulate textless S2ST as a unit-to-unit machine translation problem that requires a modest amount (dozens of hours) of parallel speech training data. We begin by pretraining an encoder-decoder unit language model over self-supervised speech units using monolingual speech data, followed by finetuning it for S2ST on a low-resource parallel dataset and finally performing unsupervised backtranslation to further improve performance, overall referred to as Pretrain-Finetune-Backtranslate a.k.a \texttt{PFB}. Figure~\ref{fig:intro} illustrates our method, comparing it to previous work. Modeling real speech data with speech unit sequences poses challenges, such as inherent unit sequence noise and ambiguity, that are orthogonal to our research questions. Thus, for simplicity, we use single-speaker synthesized speech data to train and evaluate our models, following early S2ST work~\citep{Jia2019DirectST}.

We train two English $\leftrightarrow$ German S2ST models in the European Parliament~\citep{https://doi.org/10.48550/arxiv.1911.03167} and Common Voice~\citep{ardila2020common} domains and two English $\leftrightarrow$ Marathi S2ST models in the European Parliament~\citep{https://doi.org/10.48550/arxiv.1911.03167} and All India Radio~\citep{bhogale2022effectiveness} domains, and evaluate the  en$\rightarrow$de, de$\rightarrow$en and mr$\rightarrow$en translation directions. We find that with just 20 hrs of parallel en$\rightarrow$de and de$\rightarrow$en data and 60 hrs of parallel en$\rightarrow$mr and mr$\rightarrow$en data, our models achievable reasonable performance on all three domains, within 1-2 ASR-BLEU of our high-resource supervised topline for the European Parliament domain for the de$\rightarrow$en and mr$\rightarrow$en directions.
We release code and model weights at \url{https://github.com/ajd12342/textless-s2st}.

\section{Methods}
\begin{figure*}
    \centering
    \includegraphics[width=\textwidth]{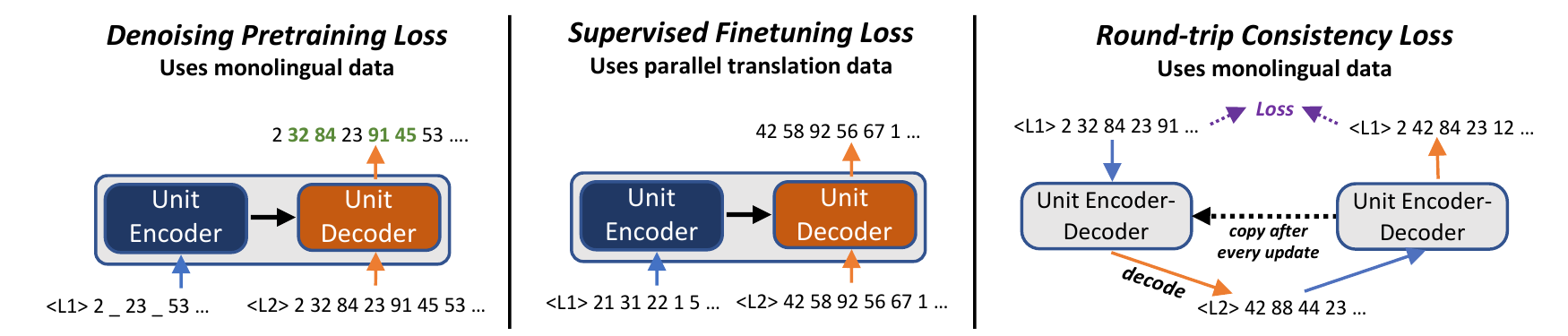}
    \caption{Training a unit-based encoder-decoder model for S2ST. The first \textbf{Pretrain} step trains on large-scale monolingual speech data using a denoising pretraining loss. The second \textbf{Finetune} step trains on low-resource parallel speech translation data using a supervised finetuning loss. The third \textbf{Backtranslate} step trains using the round-trip consistency loss (on monolingual data) and supervised finetuning replay (on parallel data).}
    \label{fig:overview}
\end{figure*}

Unsupervised S2ST~\cite{fu2023improving,wang2022simple} tackles the problem of text-based low-resource S2ST by representing input and output speech as text sequences and training a cascade of models consisting of unsupervised speech recognition~\citep{wav2vecU} (UASR), unsupervised machine translation~\citep{liu-etal-2020-multilingual-denoising} (UMT) and unsupervised text-to-speech~\citep{ni2022unsupervised} (UTTS). To adapt this for textless languages, we represent the input and output speech utterances as self-supervised discrete unit sequences rather than text sequences. Instead of UASR, we use a speech-to-unit encoder (S2U) and instead of UTTS, we use a unit-to-speech vocoder (U2S), both largely based on prior work~\citep{Hsu2021HuBERTSS, polyak21_interspeech}. Instead of text-based UMT, we train a unit encoder-decoder (U2U) S2ST model using our three-step Pretrain-Finetune-Backtranslate (\texttt{PFB}) approach illustrated in Figure~\ref{fig:overview} adapted from the unsupervised MT literature~\cite{Lample2018UnsupervisedMT}. We now describe each of these components below.%

\subsection{Speech-to-unit Encoder (S2U)}
Past work~\citep{Hsu2021HuBERTSS,chung2021w2vbert} has explored learning self-supervised discrete speech representations a.k.a speech units which preserve much of the input signal's semantic information~\citep{Pasad2021LayerWiseAO}, without needing text transcriptions to discover these units. %
It is very common to train autoregressive language models~\citep{gslm,borsos2022audiolm} over these units, enabling NLP tasks to be performed on spoken language without needing to transcribe speech waveforms into text. 

We base our speech-to-unit encoder on HuBERT~\citep{Hsu2021HuBERTSS}. We train a k-means clustering model over embeddings at an intermediate layer that maximizes the units' PNMI score, a metric that measures mutual information between phones and units. We map each embedding to its nearest k-means cluster center and apply run-length encoding~\citep{lee-etal-2022-textless}. We train a shared English-German k-means model and a separate Marathi one. We tried XLSR~\citep{conneau2020unsupervised} and Indic-wav2vec~\citep{javed2021building}, but both underperformed HuBERT when evaluated for PNMI scores. We describe training the clustering model and the evaluation of the speech-to-unit encoder in Section~\ref{subsec:s2uencoderexp}.

\subsection{Unit Encoder-Decoder (U2U)}
We train our unit encoder-decoder S2ST model using \texttt{PFB} ak.a. Pretrain-Finetune-Backtranslate (Figure~\ref{fig:overview}). We describe our general approach here and provide implementation details in Section~\ref{subsec:pfbexp}.
\paragraph{Pretrain}
We initialize with mBART-50~\citep{liu-etal-2020-multilingual-denoising} (a text encoder-decoder model), reinitializing the input/output embedding layers for our unit vocabulary. Since units can be computationally treated as text tokens with a different vocabulary, we can easily adapt the training pipeline to train on unit sequences rather than text. We pretrain using the mBART denoising objective: given a dataset $\mathcal{D}$ and a noising function $g(\cdot)$ (we use one that samples contiguous spans and masks them until a fixed ratio of tokens are masked), the decoder generates the original sequence $X$ given noised encoder input $g(X)$, optimizing model weights $\theta$ as $\argmin_{\theta}\sum_{X \in \mathcal{D}}{-\log \Pr(X | g(X); \theta)}$.%

We train two bilingual unit LMs, one for English-German, and one for English-Marathi. They are trained on unit sequences, derived from monolingual speech corpora in the three languages, generated by their respective S2U encoders. To tokenize unit sequences, we train one Sentencepiece~\citep{Kudo2018SentencePieceAS} BPE tokenizer per LM.
\paragraph{Finetune}
We perform supervised training on the pretrained unit LM using a small parallel S2ST corpus, where the input is a spoken utterance in the source language, and the target is a translated version spoken in the target language. During this finetuning process, we use the standard cross-entropy loss of the decoder generating the target unit sequence when the ground truth source unit sequence is provided to the encoder. 
\paragraph{Backtranslate}
Finally, we perform unsupervised backtranslation~\citep{Lample2018UnsupervisedMT} on our finetuned model. We follow the standard recipes used in unsupervised text backtranslation, with minor modifications to stabilize training in the speech domain. While similar to DUB~\citep{zhang2023dubdiscreteunitbacktranslation}, we perform backtranslation entirely over speech unit sequences in a textless fashion, while DUB performs backtranslation between speech and text sequences. Our backtranslation approach reconstructs a unit sequence from a model-generated synthetic translation of the same unit sequence using a round-trip translation consistency loss (visualized in Figure~\ref{fig:overview}). We start with the initial model $\mathcal{M}$ (the `backward' model) and make a copy of it, calling it $\mathcal{M'}$ (the `forward' model). Then, for every training step, we run:
\begin{enumerate}[noitemsep,nolistsep]
\item Get two batches of utterances in the two languages, $B_1$ and $B_2$.
    \item Use $\mathcal{M'}$ to translate $B_1$ to translations $B'_1$, and $B_2$ to translations $B'_2$; this step is inference only and no gradient updates occur.
    \item Given $B'_1,B'_2$ as input respectively, compute the decoder cross-entropy loss for the model $\mathcal{M}$ to reconstruct the original utterances $B_1,B_2$. Using this loss, perform a gradient update on $\mathcal{M}$'s parameters.
    \item Copy the updated parameters of $\mathcal{M}$ to $\mathcal{M'}$.
\end{enumerate}
The above corresponds to online backtranslation, where the `forward' model $\mathcal{M'}$ (generating the synthetic translation) is the same as the `backward' model $\mathcal{M}$ (used to compute the cross-entropy loss). We also explored offline backtranslation, which updates the forward model every epoch, but did not see much difference in performance. Unlike in unsupervised text backtranslation, the training was unstable in both settings. To resolve this, we mix in some supervised data (used in the finetuning step) with online backtranslation during this last stage, which stabilizes learning and shows gains.
\begin{table*}[ht!]
\centering
\setlength{\tabcolsep}{0.3em}
\begin{tabular}{ccccccc}\toprule
Model Name & Languages & Pretrain & Finetune & Backtranslate& Evaluation \\ %
\midrule
$M {\texttt{de}}^{\texttt{EP}}$ & \multirow{2}{*}{de,en} & VP (777h) + & EP-ST (20h) & VP (777h) & EP-ST (9h) en$\leftrightarrow$de\\
$M {\texttt{de}}^{\texttt{CV}}$  & & EP (5381h) & CVSS (20h) & CV (382h) & CVSS (15h) de$\rightarrow$en \\
\cmidrule{1-6}
$M {\texttt{mr}}^{\texttt{EP}}$  & \multirow{2}{*}{mr,en} & VP (529h) + & S-EP-ST (60hr) & VP (529h) + & S-EP-ST (9h) mr$\rightarrow$en \\
$M {\texttt{mr}}^{\texttt{Shr}}$ & & Shr (1000h) & S-Shr-ST (60hr) & Shr (1000h) & S-Shr-ST (10h) mr$\rightarrow$en\\
\bottomrule
\end{tabular}
\caption{Model configurations. For each dataset, we mark their duration in parentheses. Abbreviations: VP = Voxpopuli, EP = Europarl, EP-ST = Europarl-ST, CV = CommonVoice, Shr = Shrutilipi, S-EP-ST = Synth-Europarl-ST, S-Shr-ST = Synth-Shrutilipi-ST.}
\label{tab:modelsoverview}
\end{table*}
\subsection{Unit-to-speech Vocoder (U2S)}
We adapt prior work~\citep{polyak21_interspeech} on speech resynthesis from discrete units to build our unit-to-speech vocoder\footnote{\url{https://github.com/facebookresearch/speech-resynthesis/tree/main/examples/speech_to_speech_translation}}; please refer to this work for details of their approach. Given a dataset consisting of speech waveforms and their corresponding unit sequences generated by the S2U encoder, the model trains two submodules; a duration prediction module and a HiFi-GAN~\citep{kong2020hifigan} that converts unit sequences back to speech waveforms.
We train separate U2S vocoders for each language (English, German, Marathi).
\section{Experimental Setup}
\label{sec:expsetup}

\subsection{Datasets}
\label{subsec:trainingdata}

Table~\ref{tab:modelsoverview} summarizes datasets used in our work. Durations reported for parallel translation datasets correspond to durations of the source speech. More dataset details are available in Table~\ref{tab:datasets2}.

\paragraph{English-German} For pretraining, we use the union of the transcribed set of Voxpopuli~\citep{voxpopuli} and randomly-sampled subsets of the Europarl v3~\citep{koehn-2005-europarl} train set that we call Europarl-small and Europarl-mid (statistics in Table~\ref{tab:datasets2} of Appendix~\ref{app:datasets}), collected from European Parliament recordings. For finetuning, we use two datasets: (1) randomly-sampled 20-hr (10-hr per translation direction i.e. en$\rightarrow$de and de$\rightarrow$en) subset of the Europarl-ST~\citep{https://doi.org/10.48550/arxiv.1911.03167} train set and (2) randomly-sampled 20-hr (10-hr per translation direction) subset of the CVSS~\citep{jia2022cvss} train set. For the last backtranslation step, we use Voxpopuli and Common Voice 4~\citep{ardila2020common} data for the round-trip consistency loss. Common Voice and CVSS are collected using the Mozilla Common Voice project and consist of recordings of crowd-sourced workers reading out sentences primarily derived from Wikipedia; they do not belong to the European Parliament domain. For evaluation, we use Europarl-ST~\citep{https://doi.org/10.48550/arxiv.1911.03167} (for both de$\rightarrow$en and en$\rightarrow$de) and CVSS~\citep{jia2022cvss} (for de$\rightarrow$en) test sets.

\paragraph{English-Marathi}
For pretraining, we use the union of the Shrutilipi~\citep{bhogale2022effectiveness} Marathi dataset, collected from All India Radio broadcasts and the English transcribed Voxpopuli set. We were unable to find domain-matched speech translation datasets for Marathi-English. Thus, we synthetically generate parallel datasets by translating the source language utterance to target language utterance using the Google Translate API\footnote{\url{https://cloud.google.com/translate/docs/advanced/batch-translation}}. An author of this paper, who speaks both Marathi and English, manually checked a few utterances and found the translations to be of high quality. We construct two such datasets, each consisting of train and test sets: (1) Synth-Europarl-ST: translating the English side of the English-German Europarl-ST train and test sets to Marathi. (2) Synth-Shrutilipi-ST: translating 100-hr and 10-hr subsets of the Shrutilipi dataset to English, creating a train and test set respectively.

For finetuning, we randomly sampled 60-hr (30-hr per translation direction) subsets of the train sets of these two datasets. We empirically found that we need more data in English-Marathi compared to English-German, which we hypothesize is due to greater language and domain dissimilarities. For the backtranslation step, we use the union of Voxpopuli and Shrutilipi datasets for the round-trip consistency loss. For evaluation, we use the test sets of these Synth-Europarl-ST (where Marathi is translated from English), and Synth-Shrutilipi-ST datasets, (where English is translated from Marathi). We only evaluate the mr$\rightarrow$en translation direction for both. None of the targets in the test sets of either dataset have been seen during pretraining, making them suitable for use. 

\subsection{Model Configurations}
\label{subsec:modelconfigs}

Table~\ref{tab:modelsoverview} describes training and evaluation datasets for each of our four models. $M {\texttt{de}}^{\texttt{EP}}$ is trained and evaluated entirely within the European Parliament domain: it is pretrained on the union of Voxpopuli and Europarl v3, finetuned on Europarl-ST, backtranslated with Voxpopuli, and evaluated on Europarl-ST. $M {\texttt{de}}^{\texttt{CV}}$ uses the same pretraining, but is finetuned on CVSS, backtranslated with Common Voice 4.0, and evaluated on CVSS. Common Voice and CVSS consist of crowd-sourced speech recordings reading aloud sentences primarily derived from Wikipedia, which differ from the European Parliament domain. $M {\texttt{mr}}^{\texttt{EP}}$ and $M {\texttt{mr}}^{\texttt{Shr}}$ are both pretrained and backtranslated with the union of Voxpopuli and Shrutilipi i.e. English European Parliament data and Marathi All India Radio data. $M {\texttt{mr}}^{\texttt{EP}}$ is finetuned and evaluated on the European Parliament domain using Synth-Europarl-ST while $M {\texttt{mr}}^{\texttt{Shr}}$ is finetuned and evaluated on the All India Radio domain using Synth-Shrutilipi-ST.

\subsection{Generating Synthetic Speech Data} 
\label{subsec:genspeechdata}
We use single-speaker synthesized speech data for both training and evaluation, following early S2ST work~\citep{Jia2019DirectST}. All of our training datasets have ground truth transcripts; thus, we use TTS models to regenerate the speech from them and use the synthesized speech.
We use Coqui-AI's TTS software for English and German.\footnote{We use the \texttt{en/ljspeech/vits} model for English and \texttt{de/thorsten/vits} model for German. \url{https://github.com/coqui-ai/TTS})} These are VITS~\citep{kim2021conditional} models, 
trained on LJSpeech~\citep{ljspeech17} and Thorsten~\citep{Muller_Thorsten-Voice}; each have $24$ hrs of clean read speech. We use IndicTTS~\citep{kumar2023building} for Marathi; this is a FastPitch~\citep{lancucki2021fastpitch} model trained on the IndicTTS Database~\citep{baby2016resources} that contains around $3$ hrs of clean read speech. %

\section{Model Implementation}
\subsection{Speech-to-Unit Encoder (S2U)}
\label{subsec:s2uencoderexp}
To choose the speech encoder model and optimal layer, we compare the unit-phoneme PNMI scores of different choices. We decide upon using HuBERT~\citep{Hsu2021HuBERTSS}, with a shared English-German k-means model (with 200 clusters) and a standalone Marathi k-means model (with 100 clusters). We use the 6th HuBERT layer for English and German and the 8th HuBERT layer for Marathi; more details in Appendix~\ref{app:s2uresults}.  
\subsection{Unit Encoder-Decoder (U2U)}
\label{subsec:pfbexp}

\paragraph{Preprocessing} We train one Sentencepiece BPE tokenizer per LM on speech units with a $10000$-size vocab, using Voxpopuli for English-German and Voxpopuli plus Shrutilipi for English-Marathi.
\paragraph{Pretrain}%
Both LMs are initialized with \texttt{mbart-large-50} ~\citep{liu-etal-2020-multilingual-denoising}; we reinitialize input/output embedding layers. The noising function $g$ is similar to mBART; until $35\%$ masked tokens, we sample a span length $l$ from a mean-$\lambda$ Poisson distribution and replace a random contiguous sequence of length $l$ with a MASK token. For English-German model, we pretrain it in several stages with increasing task difficulty. We first train on Voxpopuli for 900k steps with $\lambda=2$. Then, we train on Voxpopuli plus Europarl-small for 5400k steps (2700k with $\lambda=2$ and 2700k with $\lambda=8$). We finally train on Voxpopuli plus Europarl-mid for 2700k steps. For English-Marathi, we train on Voxpopuli plus Shrutilipi with $\lambda=2$ for 900k steps.
 
For both LMs, the LR scheduler starts with $10^{-7}$, linearly warms up to $10^{-5}$, and then exponentially decays to $10^{-6}$. We train on 4 NVIDIA A40 GPUs with a batch size of $3125$ tokens per language for English-German and $6250$ tokens per language for English-Marathi.
\paragraph{Finetune}
We use label smoothing, dropout of $0.2$ and a learning rate of $3 \times 10^{-5}$. We train for $40$ epochs with a total batch size of $3748$ tokens on 4 GPUs. We finetune all parameters of the models except for $M{\texttt{de}}^{\texttt{EP}}$, for which we finetune only the last $5$ layers of both encoder and decoder as it shows performance gains.
\paragraph{Backtranslate}
When sampling forward translations, we use nucleus sampling~\citep{Holtzman2019TheCC} with top-p value of $0.9$ and the temperature of $0.5$. We use label smoothing of $0.2$, learning rate of 3e-5 and train for $3$ epochs with a total batch size of $3748$ tokens on 4 GPUs.
\subsection{Unit-to-Speech Vocoder (U2S)}
A separate vocoder is trained for each language, mapping from the unit vocabulary (size 200 for English-German, size 100 for Marathi) to speech clips at 16kHz, trained on the (speech, unit sequence) pairs generated by the S2U encoder, largely following~\citet{polyak21_interspeech}. We evaluate the resynthesis quality of cascading S2U+U2S in Appendix~\ref{app:resynthesis}.

\begin{table*}[ht!]
\centering
\setlength{\tabcolsep}{2pt}
\begin{tabular}{rlcccc}
\toprule
& & & \multicolumn{3}{c}{\textbf{ASR-BLEU} $\uparrow$} \\
\cmidrule(lr){4-6}
& & & \multicolumn{2}{c}{\textbf{Europarl-ST}} & \textbf{CVSS} \\
\cmidrule(lr){4-5} \cmidrule(lr){6-6}
& \textbf{Model} & \textbf{Parallel} \#\textbf{hrs} & \textbf{de$\rightarrow$en} & \textbf{en$\rightarrow$de} & \textbf{de$\rightarrow$en} \\
\midrule
& \textbf{\small{Topline models}} & & & & \\
& \textit{\small{Text-based Parallel-Low-Resource S2ST}} & & & & \\
\circled{a} & ASR $\rightarrow$ MT $\rightarrow$ TTS~(Section~\ref{subsec:comparison}) & 20h & 23.7 & 21.3 & - \\
\circled{b} & UASR $\rightarrow$ UMT $\rightarrow$ UTTS  ~\citep{fu2023improving} & 0h & - & - & 14.7 \\
\cdashlinelr{1-6}
& \textit{\small{Textless Parallel-High-Resource S2ST}} & & & & \\
\circled{c} & Bilingual S2S~\citep{duquenne2022speechmatrix} & $\approx$2600h & 16.3 & 10.1 & - \\
\circled{d} & Multilingual UTUT~\citep{kim2023manytomany} & 650h~\tablefootnote{In addition to 650h of parallel German-English data, UTUT is trained on X-to-English translation data from $18$ other languages, totaling $\approx 12000$ hours of parallel data.} & 15.8 & 9.8 & - \\
\circled{e} & Pretrain + Full Finetune (Ours) & 110h|180h & 12.0 & 13.4 & 13.6 \\
\midrule
& \textit{\small{Textless Parallel-Low-Resource S2ST}} & & & & \\
\circled{f} & Pretrain + Finetune (Ours) & 20h & 7.8 & 6.8 & 5.8 \\
\circled{g} & + Backtranslate (Ours) & 20h & 10.0 & 8.3 & 7.7 \\
\midrule
& \textbf{\small{Ablations}} & & & & \\
\circled{h} & Text mBART + Finetune & 20h & 1.0 & 0.3 & - \\
\circled{i} & + Backtranslate & 20h & 1.3 & 0.4 & - \\
\cdashlinelr{1-6}
\circled{j} & Pretrain + Backtranslate & 0h & 0.4 & 0.1 & - \\
\cdashlinelr{1-6}
\circled{k}& Pretrain + Finetune + Backtranslate w/o replay & 20h & 4.3 & 4.0 & - \\ \bottomrule
\end{tabular}
\caption{English-German S2ST evaluation using ASR-BLEU on Europarl-ST~\citep{https://doi.org/10.48550/arxiv.1911.03167} and CVSS~\citep{jia2022cvss} test sets; higher is better. Topline models use more resources by either needing high-resource parallel data or being text-based (Section~\ref{sec:result}). The Parallel \#hrs column denotes the size of parallel translation training data. In \circled{h} it denotes that 110h of EP-ST data and 180h of CVSS data is used to train two separate toplines.}
\label{tab:resultsfinal}
\end{table*}

\section{Results}
\label{sec:result}
\subsection{Evaluation Setup}
\label{subsec:evalsetup}
We evaluate the semantics of the speech translation (whether it preserves the input speech meaning) and leave non-content aspects like naturalness to future work. We use the ASR-BLEU metric following prior work~\citep{lee-etal-2022-direct,lee-etal-2022-textless}: the BLEU between the ASR transcript of the hypothesis speech translation and the ground truth text translation. We evaluate the de$\rightarrow$en, en$\rightarrow$de and mr$\rightarrow$en language directions. We opted to not evaluate the en$\rightarrow$mr direction due to poor Marathi ASR models that resulted in excessively noisy ASR-BLEU scores.
When evaluating on Europarl-ST dataset, we use wav2vec2.0 based ASR models with greedy decoding (Huggingface models \texttt{facebook/wav2vec2-large-960h-lv60-self}, \texttt{jonatasgrosman/wav2vec2-xls-r-1b-german}) used by prior S2ST work on Europarl-ST (\citet{duquenne2022speechmatrix,wang2022simple} and others). When evaluating on CVSS dataset, we use a medium-sized Whisper ASR model used by prior S2ST work on CVSS~\citep{fu2023improving}. When evaluating Marathi-English translation, we use \texttt{facebook/wav2vec2-large-960h-lv60-self}. For computing BLEU, we use SacreBLEU with default parameters.  We generate translations from our models using beam search decoding with a beam size of $10$.

\subsection{Comparison Systems}
\label{subsec:comparison}
We categorize S2ST models based on whether they leverage text as an intermediate step or not (text-based or textless) and how much parallel translation data they use (parallel-high-resource or parallel-low-resource). Our models belong to the textless, parallel-low-resource setting. Due to the lack of baselines in this setting, we instead contrast our models with existing \textbf{topline models} trained with more resources, which serve as upper bounds:

\textbf{Text-based Parallel-Low-Resource S2ST models:} 
\circled{a} is a cascaded ASR $\rightarrow$ MT $\rightarrow$ TTS system where the MT model is text mBART finetuned on the transcripts of the 20-hr low-resource parallel speech data used by our models. We use the ASR systems used for computing ASR-BLEU (Section~\ref{subsec:evalsetup}) and the TTS systems used for generating our data (Section~\ref{subsec:genspeechdata}). 
\circled{b}~\citep{fu2023improving} uses a cascaded unsupervised ASR - unsupervised MT - unsupervised TTS model that is trained on large amounts of monolingual speech and text data. %

\textbf{Textless Parallel-High-Resource S2ST models}: \circled{c} is a bilingual S2ST model trained on a large, mined SpeechMatrix dataset ($\approx$ 2600 hrs of source speech for the en$\rightarrow$de and the de$\rightarrow$en directions combined) by~\citet{duquenne2022speechmatrix}.
\circled{d}~\citep{kim2023manytomany} is a multilingual S2ST model trained on 650h of parallel aligned English-German Voxpopuli data, and about 12k hours of parallel aligned data in 18 other X-to-English language pairs. \circled{e} and \circled{l} are our pretrained unit LMs fine-tuned on more data than our parallel-low-resource models i.e. the Europarl-ST train set (110 hours), the CVSS train set (180 hours), the Synth-Europarl-ST train set (125h) and the Synth-Shrutilipi-ST train set (176h) using the same hyperparameters as our four parallel-low-resource models.

Our \textbf{Textless Parallel-Low-Resource S2ST models} consist of four models trained on different domains: $M {\texttt{de}}^{\texttt{EP}}$,$M {\texttt{de}}^{\texttt{CV}}$,$M {\texttt{mr}}^{\texttt{EP}}$ and $M {\texttt{mr}}^{\texttt{Shr}}$ as described in Section~\ref{subsec:modelconfigs}.
We evaluate each model with its in-domain evaluation data, i.e., $M {\texttt{de}}^{\texttt{EP}}$ model on Europarl-ST, $M {\texttt{de}}^{\texttt{CV}}$ model on CVSS, $M {\texttt{mr}}^{\texttt{EP}}$ on Synth-Europarl-ST, and the $M {\texttt{mr}}^{\texttt{Shr}}$ model on Synth-Shrutilipi-ST. \circled{f} and \circled{m} report the model performance after our pretraining and finetuning steps. \circled{g} and \circled{n} report the model performance after performing backtranslation. 

\subsection{Main Results}
\begin{table}[t]
\centering
\setlength{\tabcolsep}{0.8pt}
\begin{tabular}{rlccc}
\toprule
& & & \multicolumn{2}{c}{\textbf{ASR-BLEU} $\uparrow$} \\
\cmidrule(lr){4-5}
& & & \textbf{EP-ST } & \textbf{Shr-ST} \\
& \textbf{Model} & \textbf{Par.} \#\textbf{hrs} & \multicolumn{2}{c}{\textbf{mr$\rightarrow$en}} \\
\midrule
& \textbf{\small{Topline models}} & & & \\
& \textit{\small{Textless Par.-High-Res.}} & & & \\
\circled{l} & Full FT (Ours) & 125|176h & 10.9 & 17.8 \\
\midrule
& \textit{\small{Textless Par.-Low-Res.}} & & & \\
\circled{m} & Pretrain + FT (Ours) & 60h & 8.3 & 9.6 \\
\circled{n} & + BackT (Ours) & 60h & 9.2 & 10.0 \\
\bottomrule
\end{tabular}
\caption{Marathi-English S2ST evaluation using ASR-BLEU on Synth-Europarl-ST and Synth-Shrutilipi-ST test sets; higher is better. The Par. \#hrs column denotes the size of parallel training data. In \circled{o} it denotes that 125h of Synth-Europarl-ST data and 176h of Synth-Shrutilipi-ST data is used to train two separate toplines.}
\label{tab:resultsfinalmarathi}
\end{table}

We present results for the English-German pair in Table~\ref{tab:resultsfinal} and the English-Marathi pair in Table~\ref{tab:resultsfinalmarathi}. We first observe that the text-based parallel-low-resource S2ST topline models (\circled{a}-\circled{b}) trained with at most 20 hrs of parallel data outperform the best textless S2ST topline models trained with far more parallel speech data (\circled{c}-\circled{e}). This underscores the inherent task difficulty of learning purely textless S2ST models in the speech domain, even with access to far more training data. 

Next, we discuss our textless parallel-low-resource models (rows \circled{f}, \circled{g}, \circled{m} and \circled{n}). Rows \circled{f} and \circled{m} show that our models, given only 20 hr of parallel data (for English-German) and 60 hr of parallel data (for English-Marathi), learn S2ST models with reasonable BLEU scores which consistently improve post-backtranslation in rows \circled{g} and \circled{n}. Our de$\rightarrow$en Europarl-ST and the mr$\rightarrow$en Synth-Europarl-ST models are even within 1-2 BLEU of our supervised toplines \circled{e} and \circled{l} despite being trained on much less data. Another observation is regarding domain effects: the gap between our textless low-resource models and the textless high-resource toplines is smaller for European Parliament domain as compared to the Common Voice and All India Radio domains, likely due to pretrain-finetune domain mismatch as during pretraining, the models only ever see European Parliament domain English data. Finally, a qualitative analysis, based on manually looking at example outputs in Appendix~\ref{app:examples} shows that our models often preserve the semantics of the input utterance, but make egregious grammatical and language modeling mistakes.

Overall, while some of our models show encouraging results close to supervised toplines in the European Parliament domain, they underperform text-based and textless high-resource toplines.

\subsection{Ablations}
We perform ablations on the $M {\texttt{de}}^{\texttt{EP}}$ model.
\paragraph{Ablating pretraining} Our LM is initialized from the text mBART checkpoint, and then trained on a unit-based denoising objective. Without this pretraining (i.e., finetuning and backtranslating with the base mBART checkpoint), as seen in rows \circled{h} and \circled{i}, we obtain very low ASR-BLEUs less than 2 points. These results suggest that unit LM pretraining is essential in order to learn good S2ST systems in parallel-low-resource settings.

\paragraph{Ablating finetuning} We finetune the pretrained unit LM with the backtranslation round-trip consistency loss without first finetuning with parallel data. The result, \circled{j}, shows that this does not work, with near-zero BLEU scores. This suggest some amount of parallel speech is necessary.

\paragraph{Ablating replay in backtranslation}
We have already seen that adding backtranslation after finetuning boosts performance by 1-2 BLEU; compare row \circled{f} to \circled{g} or row \circled{m} to \circled{n}. We replay the supervised low-resource parallel finetuning data during backtranslation to stabilize training. We ablate training with this replay by running the backtranslation step with just the round-trip consistency loss. The result, row \circled{k}, shows that the performance worsens compared to the initialization of row \circled{f}. With replay, however, we get the results in row \circled{g}, which improve upon the initialization.

\section{Related Work}
\subsection{Speech-to-Speech Translation (S2ST)}
While cascaded S2ST models~\citep{1597243,Wahlster2000VerbmobilFO} with intermediate text translations have existed for a long time, end-to-end S2ST models start with~\citet{Jia2019DirectST}, a model that directly translates source language speech waveforms to speech waveforms in the target language. Several S2ST models~\citep{Jia2019DirectST,translatotron2,lee-etal-2022-direct,https://doi.org/10.48550/arxiv.2212.08055} are text-based i.e. they use textual supervision to stabilize training or improve performance, while other S2ST models~\citep{lee-etal-2022-textless,li2022textless,kim2023manytomany, zhu2023diffs2ut,zhang2020uwspeechspeechspeechtranslation} are textless, usually by representing speech using phonemes (which require some level of linguistic supervision) or self-supervised speech units. Most textless S2ST models require large training datasets of parallel speech translation data. One exception is work on S2ST for the textless language Hokkien~\citep{chen2022speechtospeechtranslationrealworldunwritten}; while it does not require parallel speech translation data and supports a textless language, it does require the use of a text-based pivot language (Mandarin Chinese) and large-scale parallel text data involving this pivot language. Sentence-embedding based approaches like T-Modules~\citep{Duquenne2022TModulesTM} and SONAR~\citep{duquenne2023sonarsentencelevelmultimodallanguageagnostic} circumvent the need for parallel speech translation data but still require parallel text translation data to construct strong cross-lingual embedding spaces.

In order to reduce this dependency on parallel data, unsupervised S2ST systems~\citep{wang2022simple,fu2023improving,nachmani2023translatotron} that do not use any parallel data at all have been recently proposed. However, none of them are textless; they either train cascaded S2ST models (ASR$\rightarrow$MT$\rightarrow$TTS) using unsupervised ASR~\citep{liu2022endtoend}, unsupervised MT~\citep{liu-etal-2020-multilingual-denoising} and unsupervised TTS~\citep{liu2022_unsup_tts}, or use text during training~\citep{nachmani2023translatotron}. Thus, the crucial cross-lingual translation component is learned over text tokens, limiting applicability to spoken languages. Our textless, parallel-low-resource S2ST model aims to bridge these camps.

\subsection{Large-Scale Speech-Text Models}
Several large-scale speech-text models~\citep{rubenstein2023audiopalmlargelanguagemodel,dong2023polyvoicelanguagemodelsspeech} excel at multiple speech tasks, including speech-to-speech translation. These models are trained on large-scale monolingual speech and text data as well as speech recognition, machine translation, speech translation and text-to-speech data; while they do not fit in a textless, low-resource data regime, they serve as foundation models that can be potentially used to extend to new textless, low-resource languages.

\subsection{Textless and Unit-Based NLP}
While we tackle textless S2ST, textless speech processing has studied in other tasks such as spoken language modeling~\citep{borsos2022audiolm,gslm,hassid2024textually}, emotion conversion~\citep{textless_emotion_conversion}, image-speech retrieval~\citep{Harwath2016UnsupervisedLO,peng2021}, spoken question answering~\citep{lin2022dual} and speech evaluation~\citep{https://doi.org/10.48550/arxiv.2212.08486,besacier2023textless}. Furthermore, progress in several other speech tasks like TTS~\citep{wang2023neural} that involve both speech and text has been achieved by using powerful self-supervised units (semantic units like HuBERT~\citep{Hsu2021HuBERTSS} and acoustic units like EnCodec~\citep{défossez2022high}).

\section{Conclusion}
We present the first textless low-resource speech-to-speech translation system, capable of learning from dozens of hours of parallel translation data, built by pretraining, finetuning, and backtranslating a language model over self-supervised speech unit sequences rather than text. We demonstrate its efficacy on 2 language pairs (English-German and English-Marathi) across 3 different domains. While our models achieve a decent translation performance, close to supervised toplines in some cases, they still underperform models trained on far more data or models that make use of text data, implying that several challenges still remain to make these models highly performant. However, our approach holds great promise for modelling low-resource, primarily spoken languages. We hypothesize, based on similar findings for text machine translation, that scaling our approach to a larger unit LM pretrained on more data will improve performance and may unlock unsupervised textless S2ST akin to unsupervised text MT~\citep{liu-etal-2020-multilingual-denoising}. Future work can investigate use of better S2U unit encoders for training better unit LMs, and training unit LMs on a larger set of languages.

\section*{Limitations}
Textless S2ST models, including ours, still lag in performance behind their text-based counterparts. Therefore, while they work for all languages in theory, they are currently useful only for fully textless languages and should not be used in cases where text data is readily available. Strong open-source pretrained multilingual unit language models are as yet unavailable; as a consequence, the unit LMs we use via our own pretraining have been trained on our limited compute budget and cannot yet benefit from the scale of modern text-based LLMs. Our models are trained and evaluated on synthesized single-speaker data, following early S2ST work. They do not fully generalize to real speech data that has noise and unseen speakers.

\bibliography{references}

\begin{thebibliography}{66}
\providecommand{\natexlab}[1]{#1}

\bibitem[{Ardila et~al.(2020)Ardila, Branson, Davis, Henretty, Kohler, Meyer, Morais, Saunders, Tyers, and Weber}]{ardila2020common}
Rosana Ardila, Megan Branson, Kelly Davis, Michael Henretty, Michael Kohler, Josh Meyer, Reuben Morais, Lindsay Saunders, Francis~M. Tyers, and Gregor Weber. 2020.
\newblock \href {https://arxiv.org/abs/1912.06670} {Common voice: A massively-multilingual speech corpus}.
\newblock \emph{Preprint}, arXiv:1912.06670.

\bibitem[{Baby et~al.(2016)Baby, Thomas, Nishanthi, Consortium et~al.}]{baby2016resources}
Arun Baby, Anju~Leela Thomas, NL~Nishanthi, TTS Consortium, et~al. 2016.
\newblock Resources for indian languages.
\newblock In \emph{Proceedings of Text, Speech and Dialogue}.

\bibitem[{Baevski et~al.(2021)Baevski, Hsu, CONNEAU, and Auli}]{wav2vecU}
Alexei Baevski, Wei-Ning Hsu, Alexis CONNEAU, and Michael Auli. 2021.
\newblock Unsupervised speech recognition.
\newblock In \emph{Advances in Neural Information Processing Systems}, volume~34.

\bibitem[{Besacier et~al.(2023)Besacier, Ribeiro, Galibert, and Calapodescu}]{besacier2023textless}
Laurent Besacier, Swen Ribeiro, Olivier Galibert, and Ioan Calapodescu. 2023.
\newblock \href {https://arxiv.org/abs/2210.11835} {A textless metric for speech-to-speech comparison}.
\newblock \emph{Preprint}, arXiv:2210.11835.

\bibitem[{Bhogale et~al.(2022)Bhogale, Raman, Javed, Doddapaneni, Kunchukuttan, Kumar, and Khapra}]{bhogale2022effectiveness}
Kaushal~Santosh Bhogale, Abhigyan Raman, Tahir Javed, Sumanth Doddapaneni, Anoop Kunchukuttan, Pratyush Kumar, and Mitesh~M. Khapra. 2022.
\newblock \href {https://arxiv.org/abs/2208.12666} {Effectiveness of mining audio and text pairs from public data for improving asr systems for low-resource languages}.
\newblock \emph{Preprint}, arXiv:2208.12666.

\bibitem[{Borsos et~al.(2022)Borsos, Marinier, Vincent, Kharitonov, Pietquin, Sharifi, Teboul, Grangier, Tagliasacchi, and Zeghidour}]{borsos2022audiolm}
Zalán Borsos, Raphaël Marinier, Damien Vincent, Eugene Kharitonov, Olivier Pietquin, Matt Sharifi, Olivier Teboul, David Grangier, Marco Tagliasacchi, and Neil Zeghidour. 2022.
\newblock \href {https://doi.org/10.48550/ARXIV.2209.03143} {Audiolm: a language modeling approach to audio generation}.
\newblock \emph{arXiv preprint}.

\bibitem[{Chen et~al.(2022{\natexlab{a}})Chen, Duquenne, Andrews, Kao, Mourachko, Schwenk, and Costa-jussà}]{https://doi.org/10.48550/arxiv.2212.08486}
Mingda Chen, Paul-Ambroise Duquenne, Pierre Andrews, Justine Kao, Alexandre Mourachko, Holger Schwenk, and Marta~R. Costa-jussà. 2022{\natexlab{a}}.
\newblock \href {https://doi.org/10.48550/ARXIV.2212.08486} {Blaser: A text-free speech-to-speech translation evaluation metric}.
\newblock \emph{arXiv preprint}.

\bibitem[{Chen et~al.(2022{\natexlab{b}})Chen, Tran, Yang, Du, Kao, Chung, Tomasello, Duquenne, Schwenk, Gong, Inaguma, Popuri, Wang, Pino, Hsu, and Lee}]{chen2022speechtospeechtranslationrealworldunwritten}
Peng-Jen Chen, Kevin Tran, Yilin Yang, Jingfei Du, Justine Kao, Yu-An Chung, Paden Tomasello, Paul-Ambroise Duquenne, Holger Schwenk, Hongyu Gong, Hirofumi Inaguma, Sravya Popuri, Changhan Wang, Juan Pino, Wei-Ning Hsu, and Ann Lee. 2022{\natexlab{b}}.
\newblock \href {https://arxiv.org/abs/2211.06474} {Speech-to-speech translation for a real-world unwritten language}.
\newblock \emph{Preprint}, arXiv:2211.06474.

\bibitem[{Chung et~al.(2021)Chung, Zhang, Han, Chiu, Qin, Pang, and Wu}]{chung2021w2vbert}
Yu-An Chung, Yu~Zhang, Wei Han, Chung-Cheng Chiu, James Qin, Ruoming Pang, and Yonghui Wu. 2021.
\newblock \href {https://arxiv.org/abs/2108.06209} {W2v-bert: Combining contrastive learning and masked language modeling for self-supervised speech pre-training}.
\newblock \emph{Preprint}, arXiv:2108.06209.

\bibitem[{Conneau et~al.(2020)Conneau, Baevski, Collobert, Mohamed, and Auli}]{conneau2020unsupervised}
Alexis Conneau, Alexei Baevski, Ronan Collobert, Abdelrahman Mohamed, and Michael Auli. 2020.
\newblock \href {https://arxiv.org/abs/2006.13979} {Unsupervised cross-lingual representation learning for speech recognition}.
\newblock \emph{Preprint}, arXiv:2006.13979.

\bibitem[{Dong et~al.(2023)Dong, Huang, Tian, Xu, Ko, Zhao, Feng, Li, Wang, Cheng, Yue, Bai, Chen, Lu, Ma, Wang, Wang, and Wang}]{dong2023polyvoicelanguagemodelsspeech}
Qianqian Dong, Zhiying Huang, Qiao Tian, Chen Xu, Tom Ko, Yunlong Zhao, Siyuan Feng, Tang Li, Kexin Wang, Xuxin Cheng, Fengpeng Yue, Ye~Bai, Xi~Chen, Lu~Lu, Zejun Ma, Yuping Wang, Mingxuan Wang, and Yuxuan Wang. 2023.
\newblock \href {https://arxiv.org/abs/2306.02982} {Polyvoice: Language models for speech to speech translation}.
\newblock \emph{Preprint}, arXiv:2306.02982.

\bibitem[{Duquenne et~al.(2022{\natexlab{a}})Duquenne, Gong, Dong, Du, Lee, Goswani, Wang, Pino, Sagot, and Schwenk}]{duquenne2022speechmatrix}
Paul-Ambroise Duquenne, Hongyu Gong, Ning Dong, Jingfei Du, Ann Lee, Vedanuj Goswani, Changhan Wang, Juan Pino, Benoît Sagot, and Holger Schwenk. 2022{\natexlab{a}}.
\newblock \href {https://arxiv.org/abs/2211.04508} {Speechmatrix: A large-scale mined corpus of multilingual speech-to-speech translations}.
\newblock \emph{Preprint}, arXiv:2211.04508.

\bibitem[{Duquenne et~al.(2022{\natexlab{b}})Duquenne, Gong, Sagot, and Schwenk}]{Duquenne2022TModulesTM}
Paul-Ambroise Duquenne, Hongyu Gong, Beno{\^i}t Sagot, and Holger Schwenk. 2022{\natexlab{b}}.
\newblock T-modules: Translation modules for zero-shot cross-modal machine translation.
\newblock \emph{ArXiv}, abs/2205.12216.

\bibitem[{Duquenne et~al.(2023)Duquenne, Schwenk, and Sagot}]{duquenne2023sonarsentencelevelmultimodallanguageagnostic}
Paul-Ambroise Duquenne, Holger Schwenk, and Benoît Sagot. 2023.
\newblock \href {https://arxiv.org/abs/2308.11466} {Sonar: Sentence-level multimodal and language-agnostic representations}.
\newblock \emph{Preprint}, arXiv:2308.11466.

\bibitem[{Défossez et~al.(2022)Défossez, Copet, Synnaeve, and Adi}]{défossez2022high}
Alexandre Défossez, Jade Copet, Gabriel Synnaeve, and Yossi Adi. 2022.
\newblock \href {https://arxiv.org/abs/2210.13438} {High fidelity neural audio compression}.
\newblock \emph{Preprint}, arXiv:2210.13438.

\bibitem[{Fu et~al.(2023)Fu, Tseng, Shi, Li, Hsu, Watanabe, and yi~Lee}]{fu2023improving}
Yu-Kuan Fu, Liang-Hsuan Tseng, Jiatong Shi, Chen-An Li, Tsu-Yuan Hsu, Shinji Watanabe, and Hung yi~Lee. 2023.
\newblock \href {https://arxiv.org/abs/2305.07455} {Improving cascaded unsupervised speech translation with denoising back-translation}.
\newblock \emph{Preprint}, arXiv:2305.07455.

\bibitem[{Haddow et~al.(2022)Haddow, Bawden, Miceli~Barone, Helcl, and Birch}]{haddow-etal-2022-survey}
Barry Haddow, Rachel Bawden, Antonio~Valerio Miceli~Barone, Jind{\v{r}}ich Helcl, and Alexandra Birch. 2022.
\newblock \href {https://doi.org/10.1162/coli_a_00446} {Survey of low-resource machine translation}.
\newblock \emph{Computational Linguistics}, 48(3):673--732.

\bibitem[{Harwath et~al.(2016)Harwath, Torralba, and Glass}]{Harwath2016UnsupervisedLO}
David~F. Harwath, A.~Torralba, and James~R. Glass. 2016.
\newblock Unsupervised learning of spoken language with visual context.
\newblock In \emph{NIPS}.

\bibitem[{Hassid et~al.(2024)Hassid, Remez, Nguyen, Gat, Conneau, Kreuk, Copet, Defossez, Synnaeve, Dupoux, Schwartz, and Adi}]{hassid2024textually}
Michael Hassid, Tal Remez, Tu~Anh Nguyen, Itai Gat, Alexis Conneau, Felix Kreuk, Jade Copet, Alexandre Defossez, Gabriel Synnaeve, Emmanuel Dupoux, Roy Schwartz, and Yossi Adi. 2024.
\newblock \href {https://arxiv.org/abs/2305.13009} {Textually pretrained speech language models}.
\newblock \emph{Preprint}, arXiv:2305.13009.

\bibitem[{Holtzman et~al.(2019)Holtzman, Buys, Forbes, and Choi}]{Holtzman2019TheCC}
Ari Holtzman, Jan Buys, Maxwell Forbes, and Yejin Choi. 2019.
\newblock The curious case of neural text degeneration.
\newblock \emph{ArXiv}, abs/1904.09751.

\bibitem[{Hsu et~al.(2021)Hsu, Bolte, Tsai, Lakhotia, Salakhutdinov, and Mohamed}]{Hsu2021HuBERTSS}
Wei-Ning Hsu, Benjamin Bolte, Yao-Hung~Hubert Tsai, Kushal Lakhotia, Ruslan Salakhutdinov, and Abdelrahman Mohamed. 2021.
\newblock Hubert: Self-supervised speech representation learning by masked prediction of hidden units.
\newblock \emph{IEEE/ACM Transactions on Audio, Speech, and Language Processing}, 29:3451--3460.

\bibitem[{Inaguma et~al.(2022)Inaguma, Popuri, Kulikov, Chen, Wang, Chung, Tang, Lee, Watanabe, and Pino}]{https://doi.org/10.48550/arxiv.2212.08055}
Hirofumi Inaguma, Sravya Popuri, Ilia Kulikov, Peng-Jen Chen, Changhan Wang, Yu-An Chung, Yun Tang, Ann Lee, Shinji Watanabe, and Juan Pino. 2022.
\newblock \href {https://doi.org/10.48550/ARXIV.2212.08055} {Unity: Two-pass direct speech-to-speech translation with discrete units}.
\newblock \emph{arXiv preprint}.

\bibitem[{Iranzo-Sánchez et~al.(2019)Iranzo-Sánchez, Silvestre-Cerdà, Jorge, Roselló, Giménez, Sanchis, Civera, and Juan}]{https://doi.org/10.48550/arxiv.1911.03167}
Javier Iranzo-Sánchez, Joan~Albert Silvestre-Cerdà, Javier Jorge, Nahuel Roselló, Adrià Giménez, Albert Sanchis, Jorge Civera, and Alfons Juan. 2019.
\newblock \href {https://doi.org/10.48550/ARXIV.1911.03167} {Europarl-st: A multilingual corpus for speech translation of parliamentary debates}.
\newblock \emph{arXiv preprint}.

\bibitem[{Ito and Johnson(2017)}]{ljspeech17}
Keith Ito and Linda Johnson. 2017.
\newblock The lj speech dataset.
\newblock \url{https://keithito.com/LJ-Speech-Dataset/}.

\bibitem[{Javed et~al.(2021)Javed, Doddapaneni, Raman, Bhogale, Ramesh, Kunchukuttan, Kumar, and Khapra}]{javed2021building}
Tahir Javed, Sumanth Doddapaneni, Abhigyan Raman, Kaushal~Santosh Bhogale, Gowtham Ramesh, Anoop Kunchukuttan, Pratyush Kumar, and Mitesh~M. Khapra. 2021.
\newblock \href {https://arxiv.org/abs/2111.03945} {Towards building asr systems for the next billion users}.
\newblock \emph{Preprint}, arXiv:2111.03945.

\bibitem[{Jia et~al.(2021)Jia, Ramanovich, Remez, and Pomerantz}]{translatotron2}
Ye~Jia, Michelle~Tadmor Ramanovich, Tal Remez, and Roi Pomerantz. 2021.
\newblock Translatotron 2: High-quality direct speech-to-speech translation with voice preservation.
\newblock \emph{arXiv preprint}.

\bibitem[{Jia et~al.(2022)Jia, Tadmor~Ramanovich, Wang, and Zen}]{jia2022cvss}
Ye~Jia, Michelle Tadmor~Ramanovich, Quan Wang, and Heiga Zen. 2022.
\newblock {CVSS} corpus and massively multilingual speech-to-speech translation.
\newblock In \emph{Proceedings of Language Resources and Evaluation Conference (LREC)}, pages 6691--6703.

\bibitem[{Jia et~al.(2019)Jia, Weiss, Biadsy, Macherey, Johnson, Chen, and Wu}]{Jia2019DirectST}
Ye~Jia, Ron~J. Weiss, Fadi Biadsy, Wolfgang Macherey, Melvin Johnson, Zhifeng Chen, and Yonghui Wu. 2019.
\newblock \href {https://arxiv.org/abs/1904.06037} {Direct speech-to-speech translation with a sequence-to-sequence model}.
\newblock In \emph{Interspeech}.

\bibitem[{Kim et~al.(2021)Kim, Kong, and Son}]{kim2021conditional}
Jaehyeon Kim, Jungil Kong, and Juhee Son. 2021.
\newblock \href {https://arxiv.org/abs/2106.06103} {Conditional variational autoencoder with adversarial learning for end-to-end text-to-speech}.
\newblock \emph{Preprint}, arXiv:2106.06103.

\bibitem[{Kim et~al.(2023)Kim, Choi, Kim, and Ro}]{kim2023manytomany}
Minsu Kim, Jeongsoo Choi, Dahun Kim, and Yong~Man Ro. 2023.
\newblock \href {https://arxiv.org/abs/2308.01831} {Many-to-many spoken language translation via unified speech and text representation learning with unit-to-unit translation}.
\newblock \emph{Preprint}, arXiv:2308.01831.

\bibitem[{Koehn(2005)}]{koehn-2005-europarl}
Philipp Koehn. 2005.
\newblock \href {https://aclanthology.org/2005.mtsummit-papers.11} {{E}uroparl: A parallel corpus for statistical machine translation}.
\newblock In \emph{Proceedings of Machine Translation Summit X: Papers}, pages 79--86, Phuket, Thailand.

\bibitem[{Kong et~al.(2020)Kong, Kim, and Bae}]{kong2020hifigan}
Jungil Kong, Jaehyeon Kim, and Jaekyoung Bae. 2020.
\newblock \href {https://arxiv.org/abs/2010.05646} {Hifi-gan: Generative adversarial networks for efficient and high fidelity speech synthesis}.
\newblock \emph{Preprint}, arXiv:2010.05646.

\bibitem[{Kreuk et~al.(2021)Kreuk, Polyak, Copet, Kharitonov, Nguyen, Rivière, Hsu, Mohamed, Dupoux, and Adi}]{textless_emotion_conversion}
Felix Kreuk, Adam Polyak, Jade Copet, Eugene Kharitonov, Tu-Anh Nguyen, Morgane Rivière, Wei-Ning Hsu, Abdelrahman Mohamed, Emmanuel Dupoux, and Yossi Adi. 2021.
\newblock \href {https://arxiv.org/abs/2111.07402} {Textless speech emotion conversion using discrete and decomposed representations}.
\newblock \emph{arXiv preprint}.

\bibitem[{Kudo and Richardson(2018)}]{Kudo2018SentencePieceAS}
Taku Kudo and John Richardson. 2018.
\newblock Sentencepiece: A simple and language independent subword tokenizer and detokenizer for neural text processing.
\newblock \emph{ArXiv}, abs/1808.06226.

\bibitem[{Kumar et~al.(2023)Kumar, au2, Kumar, Khapra, and Nandakumar}]{kumar2023building}
Gokul~Karthik Kumar, Praveen S~V au2, Pratyush Kumar, Mitesh~M. Khapra, and Karthik Nandakumar. 2023.
\newblock \href {https://arxiv.org/abs/2211.09536} {Towards building text-to-speech systems for the next billion users}.
\newblock \emph{Preprint}, arXiv:2211.09536.

\bibitem[{Lakhotia et~al.(2021)Lakhotia, Kharitonov, Hsu, Adi, Polyak, Bolte, Nguyen, Copet, Baevski, Mohamed, and Dupoux}]{gslm}
Kushal Lakhotia, Evgeny Kharitonov, Wei{-}Ning Hsu, Yossi Adi, Adam Polyak, Benjamin Bolte, Tu~Anh Nguyen, Jade Copet, Alexei Baevski, Adelrahman Mohamed, and Emmanuel Dupoux. 2021.
\newblock \href {https://arxiv.org/abs/2102.01192} {Generative spoken language modeling from raw audio}.
\newblock \emph{CoRR}.

\bibitem[{Lample et~al.(2018)Lample, Denoyer, and Ranzato}]{Lample2018UnsupervisedMT}
Guillaume Lample, Ludovic Denoyer, and Marc'Aurelio Ranzato. 2018.
\newblock Unsupervised machine translation using monolingual corpora only.
\newblock \emph{ArXiv}, abs/1711.00043.

\bibitem[{Lee et~al.(2022{\natexlab{a}})Lee, Chen, Wang, Gu, Popuri, Ma, Polyak, Adi, He, Tang, Pino, and Hsu}]{lee-etal-2022-direct}
Ann Lee, Peng-Jen Chen, Changhan Wang, Jiatao Gu, Sravya Popuri, Xutai Ma, Adam Polyak, Yossi Adi, Qing He, Yun Tang, Juan Pino, and Wei-Ning Hsu. 2022{\natexlab{a}}.
\newblock Direct speech-to-speech translation with discrete units.
\newblock In \emph{Proceedings of the 60th Annual Meeting of the Association for Computational Linguistics (Volume 1: Long Papers)}.

\bibitem[{Lee et~al.(2022{\natexlab{b}})Lee, Gong, Duquenne, Schwenk, Chen, Wang, Popuri, Adi, Pino, Gu, and Hsu}]{lee-etal-2022-textless}
Ann Lee, Hongyu Gong, Paul-Ambroise Duquenne, Holger Schwenk, Peng-Jen Chen, Changhan Wang, Sravya Popuri, Yossi Adi, Juan Pino, Jiatao Gu, and Wei-Ning Hsu. 2022{\natexlab{b}}.
\newblock Textless speech-to-speech translation on real data.
\newblock In \emph{Proceedings of the 2022 Conference of the North American Chapter of the Association for Computational Linguistics: Human Language Technologies}.

\bibitem[{Lewis et~al.(2016)Lewis, Simon, and Fennig}]{ethnologue}
M.~Paul Lewis, Gary~F. Simon, and Charles~D. Fennig. 2016.
\newblock \emph{Ethnologue: Languages of the World, Nineteenth edition}.
\newblock SIL International. Online version: http://www.ethnologue.com.

\bibitem[{Li et~al.(2022)Li, Jia, and Chiu}]{li2022textless}
Xinjian Li, Ye~Jia, and Chung-Cheng Chiu. 2022.
\newblock \href {https://arxiv.org/abs/2211.00115} {Textless direct speech-to-speech translation with discrete speech representation}.
\newblock \emph{Preprint}, arXiv:2211.00115.

\bibitem[{Lin et~al.(2022)Lin, Chuang, Chung, wen Yang, Chen, Dong, Li, Mohamed, yi~Lee, and shan Lee}]{lin2022dual}
Guan-Ting Lin, Yung-Sung Chuang, Ho-Lam Chung, Shu wen Yang, Hsuan-Jui Chen, Shuyan Dong, Shang-Wen Li, Abdelrahman Mohamed, Hung yi~Lee, and Lin shan Lee. 2022.
\newblock \href {https://arxiv.org/abs/2203.04911} {Dual: Discrete spoken unit adaptive learning for textless spoken question answering}.
\newblock \emph{Preprint}, arXiv:2203.04911.

\bibitem[{Liu et~al.(2022{\natexlab{a}})Liu, Lai, Hsu, Auli, Baevski, and Glass}]{liu2022_unsup_tts}
Alexander Liu, Cheng-I Lai, Wei-Ning Hsu, Michael Auli, Alexei Baevski, and James Glass. 2022{\natexlab{a}}.
\newblock Simple and effective unsupervised speech synthesis.
\newblock In \emph{INTERSPEECH}.

\bibitem[{Liu et~al.(2022{\natexlab{b}})Liu, Hsu, Auli, and Baevski}]{liu2022endtoend}
Alexander~H. Liu, Wei-Ning Hsu, Michael Auli, and Alexei Baevski. 2022{\natexlab{b}}.
\newblock \href {https://arxiv.org/abs/2204.02492} {Towards end-to-end unsupervised speech recognition}.
\newblock \emph{Preprint}, arXiv:2204.02492.

\bibitem[{Liu et~al.(2020)Liu, Gu, Goyal, Li, Edunov, Ghazvininejad, Lewis, and Zettlemoyer}]{liu-etal-2020-multilingual-denoising}
Yinhan Liu, Jiatao Gu, Naman Goyal, Xian Li, Sergey Edunov, Marjan Ghazvininejad, Mike Lewis, and Luke Zettlemoyer. 2020.
\newblock \href {https://doi.org/10.1162/tacl_a_00343} {Multilingual denoising pre-training for neural machine translation}.
\newblock \emph{Transactions of the Association for Computational Linguistics}, 8:726--742.

\bibitem[{Müller and Kreutz(2020)}]{Muller_Thorsten-Voice}
Thorsten Müller and Dominik Kreutz. 2020.
\newblock \href {https://github.com/thorstenMueller/Thorsten-Voice} {{Thorsten-Voice}}.

\bibitem[{Nachmani et~al.(2023)Nachmani, Levkovitch, Ding, Asawaroengchai, Zen, and Ramanovich}]{nachmani2023translatotron}
Eliya Nachmani, Alon Levkovitch, Yifan Ding, Chulayuth Asawaroengchai, Heiga Zen, and Michelle~Tadmor Ramanovich. 2023.
\newblock \href {https://arxiv.org/abs/2305.17547} {Translatotron 3: Speech to speech translation with monolingual data}.
\newblock \emph{Preprint}, arXiv:2305.17547.

\bibitem[{Nakamura et~al.(2006)Nakamura, Markov, Nakaiwa, Kikui, Kawai, Jitsuhiro, Zhang, Yamamoto, Sumita, and Yamamoto}]{1597243}
S.~Nakamura, K.~Markov, H.~Nakaiwa, G.~Kikui, H.~Kawai, T.~Jitsuhiro, J.-S. Zhang, H.~Yamamoto, E.~Sumita, and S.~Yamamoto. 2006.
\newblock \href {https://doi.org/10.1109/TSA.2005.860774} {The atr multilingual speech-to-speech translation system}.
\newblock \emph{IEEE Transactions on Audio, Speech, and Language Processing}, 14(2):365--376.

\bibitem[{Ni et~al.(2022)Ni, Wang, Gao, Qian, Zhang, Chang, and Hasegawa-Johnson}]{ni2022unsupervised}
Junrui Ni, Liming Wang, Heting Gao, Kaizhi Qian, Yang Zhang, Shiyu Chang, and Mark Hasegawa-Johnson. 2022.
\newblock \href {https://arxiv.org/abs/2203.15796} {Unsupervised text-to-speech synthesis by unsupervised automatic speech recognition}.
\newblock \emph{Preprint}, arXiv:2203.15796.

\bibitem[{Panayotov et~al.(2015)Panayotov, Chen, Povey, and Khudanpur}]{librispeech}
Vassil Panayotov, Guoguo Chen, Daniel Povey, and Sanjeev Khudanpur. 2015.
\newblock \href {https://doi.org/10.1109/ICASSP.2015.7178964} {Librispeech: An asr corpus based on public domain audio books}.
\newblock In \emph{2015 IEEE International Conference on Acoustics, Speech and Signal Processing (ICASSP)}, pages 5206--5210.

\bibitem[{Pasad et~al.(2021)Pasad, Chou, and Livescu}]{Pasad2021LayerWiseAO}
Ankita Pasad, Ju-Chieh Chou, and Karen Livescu. 2021.
\newblock Layer-wise analysis of a self-supervised speech representation model.
\newblock In \emph{ASRU}.

\bibitem[{Peng and Harwath(2022)}]{peng2021}
Puyuan Peng and David Harwath. 2022.
\newblock Fast-slow transformer for visually grounding speech.
\newblock In \emph{ICASSP}.

\bibitem[{Polyak et~al.(2021)Polyak, Adi, Copet, Kharitonov, Lakhotia, Hsu, Mohamed, and Dupoux}]{polyak21_interspeech}
Adam Polyak, Yossi Adi, Jade Copet, Eugene Kharitonov, Kushal Lakhotia, Wei-Ning Hsu, Abdelrahman Mohamed, and Emmanuel Dupoux. 2021.
\newblock {Speech Resynthesis from Discrete Disentangled Self-Supervised Representations}.
\newblock In \emph{Proc. Interspeech 2021}.

\bibitem[{Pratap et~al.(2020)Pratap, Xu, Sriram, Synnaeve, and Collobert}]{Pratap_2020}
Vineel Pratap, Qiantong Xu, Anuroop Sriram, Gabriel Synnaeve, and Ronan Collobert. 2020.
\newblock \href {https://doi.org/10.21437/interspeech.2020-2826} {{MLS}: A large-scale multilingual dataset for speech research}.
\newblock In \emph{Interspeech 2020}. {ISCA}.

\bibitem[{Rubenstein et~al.(2023)Rubenstein, Asawaroengchai, Nguyen, Bapna, Borsos, de~Chaumont~Quitry, Chen, Badawy, Han, Kharitonov, Muckenhirn, Padfield, Qin, Rozenberg, Sainath, Schalkwyk, Sharifi, Ramanovich, Tagliasacchi, Tudor, Velimirović, Vincent, Yu, Wang, Zayats, Zeghidour, Zhang, Zhang, Zilka, and Frank}]{rubenstein2023audiopalmlargelanguagemodel}
Paul~K. Rubenstein, Chulayuth Asawaroengchai, Duc~Dung Nguyen, Ankur Bapna, Zalán Borsos, Félix de~Chaumont~Quitry, Peter Chen, Dalia~El Badawy, Wei Han, Eugene Kharitonov, Hannah Muckenhirn, Dirk Padfield, James Qin, Danny Rozenberg, Tara Sainath, Johan Schalkwyk, Matt Sharifi, Michelle~Tadmor Ramanovich, Marco Tagliasacchi, Alexandru Tudor, Mihajlo Velimirović, Damien Vincent, Jiahui Yu, Yongqiang Wang, Vicky Zayats, Neil Zeghidour, Yu~Zhang, Zhishuai Zhang, Lukas Zilka, and Christian Frank. 2023.
\newblock \href {https://arxiv.org/abs/2306.12925} {Audiopalm: A large language model that can speak and listen}.
\newblock \emph{Preprint}, arXiv:2306.12925.

\bibitem[{{Seamless Communication} et~al.(2023){Seamless Communication}, Barrault, Chung, Meglioli, Dale, Dong, Duquenne, Elsahar, Gong, Heffernan, Hoffman, Klaiber, Li, Licht, Maillard, Rakotoarison, Sadagopan, Wenzek, Ye, Akula, Chen, Hachem, Ellis, Gonzalez, Haaheim, Hansanti, Howes, Huang, Hwang, Inaguma, Jain, Kalbassi, Kallet, Kulikov, Lam, Li, Ma, Mavlyutov, Peloquin, Ramadan, Ramakrishnan, Sun, Tran, Tran, Tufanov, Vogeti, Wood, Yang, Yu, Andrews, Balioglu, Costa-juss\`{a}, Onur, {C}elebi, Elbayad, Gao, Guzm\'an, Kao, Lee, Mourachko, Pino, Popuri, Ropers, Saleem, Schwenk, Tomasello, Wang, Wang, and Wang}]{seamlessm4t2023}
{Seamless Communication}, Lo\"{i}c Barrault, Yu-An Chung, Mariano~Cora Meglioli, David Dale, Ning Dong, Paul-Ambroise Duquenne, Hady Elsahar, Hongyu Gong, Kevin Heffernan, John Hoffman, Christopher Klaiber, Pengwei Li, Daniel Licht, Jean Maillard, Alice Rakotoarison, Kaushik~Ram Sadagopan, Guillaume Wenzek, Ethan Ye, Bapi Akula, Peng-Jen Chen, Naji~El Hachem, Brian Ellis, Gabriel~Mejia Gonzalez, Justin Haaheim, Prangthip Hansanti, Russ Howes, Bernie Huang, Min-Jae Hwang, Hirofumi Inaguma, Somya Jain, Elahe Kalbassi, Amanda Kallet, Ilia Kulikov, Janice Lam, Daniel Li, Xutai Ma, Ruslan Mavlyutov, Benjamin Peloquin, Mohamed Ramadan, Abinesh Ramakrishnan, Anna Sun, Kevin Tran, Tuan Tran, Igor Tufanov, Vish Vogeti, Carleigh Wood, Yilin Yang, Bokai Yu, Pierre Andrews, Can Balioglu, Marta~R. Costa-juss\`{a}, Onur, {C}elebi, Maha Elbayad, Cynthia Gao, Francisco Guzm\'an, Justine Kao, Ann Lee, Alexandre Mourachko, Juan Pino, Sravya Popuri, Christophe Ropers, Safiyyah Saleem, Holger Schwenk, Paden Tomasello, Changhan
  Wang, Jeff Wang, and Skyler Wang. 2023.
\newblock {SeamlessM4T—Massively Multilingual \& Multimodal Machine Translation}.
\newblock \emph{ArXiv}.

\bibitem[{Tiedemann(2012)}]{tiedemann-2012-parallel}
J{\"o}rg Tiedemann. 2012.
\newblock \href {http://www.lrec-conf.org/proceedings/lrec2012/pdf/463_Paper.pdf} {Parallel data, tools and interfaces in {OPUS}}.
\newblock In \emph{Proceedings of the Eighth International Conference on Language Resources and Evaluation ({LREC}'12)}, pages 2214--2218, Istanbul, Turkey. European Language Resources Association (ELRA).

\bibitem[{Wahlster(2000)}]{Wahlster2000VerbmobilFO}
Wolfgang Wahlster. 2000.
\newblock \href {https://api.semanticscholar.org/CorpusID:265678893} {Verbmobil: Foundations of speech-to-speech translation}.
\newblock In \emph{Artificial Intelligence}.

\bibitem[{Wang et~al.(2022{\natexlab{a}})Wang, Inaguma, Chen, Kulikov, Tang, Hsu, Auli, and Pino}]{https://doi.org/10.48550/arxiv.2210.10191}
Changhan Wang, Hirofumi Inaguma, Peng-Jen Chen, Ilia Kulikov, Yun Tang, Wei-Ning Hsu, Michael Auli, and Juan Pino. 2022{\natexlab{a}}.
\newblock \href {https://doi.org/10.48550/ARXIV.2210.10191} {Simple and effective unsupervised speech translation}.
\newblock \emph{arXiv preprint}.

\bibitem[{Wang et~al.(2022{\natexlab{b}})Wang, Inaguma, Chen, Kulikov, Tang, Hsu, Auli, and Pino}]{wang2022simple}
Changhan Wang, Hirofumi Inaguma, Peng-Jen Chen, Ilia Kulikov, Yun Tang, Wei-Ning Hsu, Michael Auli, and Juan Pino. 2022{\natexlab{b}}.
\newblock \href {https://arxiv.org/abs/2210.10191} {{Simple and Effective Unsupervised Speech Translation}}.
\newblock \emph{Preprint}, arXiv:2210.10191.

\bibitem[{Wang et~al.(2021)Wang, Riviere, Lee, Wu, Talnikar, Haziza, Williamson, Pino, and Dupoux}]{voxpopuli}
Changhan Wang, Morgane Riviere, Ann Lee, Anne Wu, Chaitanya Talnikar, Daniel Haziza, Mary Williamson, Juan Pino, and Emmanuel Dupoux. 2021.
\newblock \href {https://doi.org/10.18653/v1/2021.acl-long.80} {{V}ox{P}opuli: A large-scale multilingual speech corpus for representation learning, semi-supervised learning and interpretation}.
\newblock In \emph{Proceedings of the 59th Annual Meeting of the Association for Computational Linguistics and the 11th International Joint Conference on Natural Language Processing (Volume 1: Long Papers)}, pages 993--1003, Online. Association for Computational Linguistics.

\bibitem[{Wang et~al.(2023)Wang, Chen, Wu, Zhang, Zhou, Liu, Chen, Liu, Wang, Li, He, Zhao, and Wei}]{wang2023neural}
Chengyi Wang, Sanyuan Chen, Yu~Wu, Ziqiang Zhang, Long Zhou, Shujie Liu, Zhuo Chen, Yanqing Liu, Huaming Wang, Jinyu Li, Lei He, Sheng Zhao, and Furu Wei. 2023.
\newblock \href {https://arxiv.org/abs/2301.02111} {Neural codec language models are zero-shot text to speech synthesizers}.
\newblock \emph{Preprint}, arXiv:2301.02111.

\bibitem[{Zhang et~al.(2020)Zhang, Tan, Ren, Qin, Zhang, and Liu}]{zhang2020uwspeechspeechspeechtranslation}
Chen Zhang, Xu~Tan, Yi~Ren, Tao Qin, Kejun Zhang, and Tie-Yan Liu. 2020.
\newblock \href {https://arxiv.org/abs/2006.07926} {Uwspeech: Speech to speech translation for unwritten languages}.
\newblock \emph{Preprint}, arXiv:2006.07926.

\bibitem[{Zhang et~al.(2023)Zhang, Ye, Ko, Wang, and Zhou}]{zhang2023dubdiscreteunitbacktranslation}
Dong Zhang, Rong Ye, Tom Ko, Mingxuan Wang, and Yaqian Zhou. 2023.
\newblock \href {https://arxiv.org/abs/2305.11411} {Dub: Discrete unit back-translation for speech translation}.
\newblock \emph{Preprint}, arXiv:2305.11411.

\bibitem[{Zhu et~al.(2023)Zhu, Gao, Zhou, Ye, and Xu}]{zhu2023diffs2ut}
Yongxin Zhu, Zhujin Gao, Xinyuan Zhou, Zhongyi Ye, and Linli Xu. 2023.
\newblock \href {https://arxiv.org/abs/2310.17570} {Diffs2ut: A semantic preserving diffusion model for textless direct speech-to-speech translation}.
\newblock \emph{Preprint}, arXiv:2310.17570.

\bibitem[{Łańcucki(2021)}]{lancucki2021fastpitch}
Adrian Łańcucki. 2021.
\newblock \href {https://arxiv.org/abs/2006.06873} {Fastpitch: Parallel text-to-speech with pitch prediction}.
\newblock \emph{Preprint}, arXiv:2006.06873.

\end{thebibliography}

\appendix

\section{Datasets}
\label{app:datasets}
We provide a summary of all the datasets used in this paper in Table~\ref{tab:datasets2}.
\begin{table*}\centering
\small
\begin{tabular}{cccc}\toprule
\textbf{Module} &\textbf{Dataset} &\textbf{Duration} &\textbf{Lang} \\\midrule
\textcolor{red}{S2U Encoder: Pretraining} & \textcolor{red}{Librispeech} &\textcolor{red}{960h} &\textcolor{red}{en} \\
\midrule
\multirow{2}{*}{S2U Encoder: k-means Clustering} &Librispeech, MLS &48h, 48h &en, de \\
&Shrutilipi &100h &mr \\
\midrule
\multirow{4}{*}{U2U Pretraining} & Voxpopuli &529h, 248h &en, de \\
&Europarl-small &811h, 975h &en, de \\
&Europarl-mid &2463h, 2918h &en, de \\
&Shrutilipi &1000h &mr \\
\midrule
\multirow{4}{*}{U2U Finetuning (Toplines)} &Europarl-ST &83h,27h &en$\rightarrow$de, de$\rightarrow$en  \\
&CVSS &91h,88h&en$\rightarrow$de, de$\rightarrow$en \\
& Synth-EP-ST &83h,42h&en$\rightarrow$mr, mr$\rightarrow$en \\
& Synth-Shr-ST &76h,100h&en$\rightarrow$mr, mr$\rightarrow$en \\
\midrule
\multirow{4}{*}{U2U Finetuning (Low-Resource)} &Europarl-ST &10h,10h &en$\rightarrow$de, de$\rightarrow$en  \\
&CVSS &10h,10h&en$\rightarrow$de, de$\rightarrow$en \\
& Synth-EP-ST &30h,30h&en$\rightarrow$mr, mr$\rightarrow$en \\
& Synth-Shr-ST &30h,30h&en$\rightarrow$mr, mr$\rightarrow$en \\
\midrule
\multirow{3}{*}{U2U Backtranslation} &Voxpopuli &529h, 248h &en, de \\
&Common Voice &294h, 89h &en, de \\
&Shrutilipi &1000h &mr \\
\midrule
\multirow{2}{*}{U2S Vocoder} &Voxpopuli &529h, 248h &en, de \\
&Shrutilipi &1000h &mr \\
\midrule
\multirow{4}{*}{Evaluation} &Europarl-ST &3h,6h &en$\rightarrow$de, de$\rightarrow$en  \\
&CVSS &15h&de$\rightarrow$en \\
& Synth-EP-ST &9h&mr$\rightarrow$en \\
& Synth-Shr-ST &10h&mr$\rightarrow$en \\
\bottomrule
\end{tabular}
\caption{Summary of datasets used to develop our system, with datasets used by base pretrained models colored \textcolor{red}{red}. Datasets in the U2U Finetune and U2U Evaluation sections are parallel translation datasets, and we report duration statistics for both translation directions separately, the duration being that of the source speech.}\label{tab:datasets2}
\end{table*}

\section{Compute Details}
We train all our models on 4 NVIDIA A40s (often using 2 GPUs with gradient accumulation of 2, or 1 GPU with gradient accumulation of 1, which is equivalent to 4 GPUs).

\begin{table}
\centering
\setlength{\tabcolsep}{2pt}
\begin{tabular}{lccccc}
\toprule
\textbf{Model} & \textbf{Test Set} & \multicolumn{4}{c}{\textbf{ASR-BLEU} $\uparrow$} \\
\cmidrule(lr){3-6}
& & \textbf{short} & \textbf{med} & \textbf{long} & \textbf{all} \\ \midrule
Row \circled{g} & EP-ST de$\rightarrow$en & 10.1 & 10.6 & 9.5 & 10.0 \\
Row \circled{g} & EP-ST en$\rightarrow$de & 9.6 & 9.0 & 7.7 & 8.3 \\
Row \circled{g} & CVSS de$\rightarrow$en & 6.5 & 8.3 & 7.7 & 7.7 \\
Row \circled{n} & S-EP-ST mr$\rightarrow$en & 10.9 & 10.1 & 8.0 & 9.2 \\
Row \circled{n} & S-Shr-ST mr$\rightarrow$en & 10.9 & 13.0 & 8.0 & 10.0 \\
\bottomrule
\end{tabular}
\caption{S2ST evaluation using ASR-BLEU, broken down by test set lengths (short, medium, long) as well as the overall ASR-BLEU (all).}
\label{tab:breakdown}
\end{table}
\section{Length-wise ASR-BLEU Breakdown}
In order to investigate how our model performance differs for short, medium and long test examples, for each test dataset (Europarl-ST, CVSS, Synth-EP-ST and Synth-Shruti-ST) we compute the character lengths of every target example and compute the 33rd and 66th percentiles of the length distribution. We call all examples with a length shorter than the 33rd percentile ‘short’, ones in between the two ‘medium’, and longer than the 66th percentile ‘long’. We then evaluate our best models, row \circled{g} (for English-German) and \circled{n} (for English-Marathi) from Tables~\ref{tab:resultsfinal} and~\ref{tab:resultsfinalmarathi} on each test data subset in Table~\ref{tab:breakdown}. We see that the model does better on short/medium utterances as compared to long utterances. The performance of the long utterances is within 1 BLEU point of the overall performance.

\begin{figure*}[ht!]
    \centering
    \begin{subfigure}{0.45\textwidth}
        \centering
        \includegraphics[width=\linewidth]{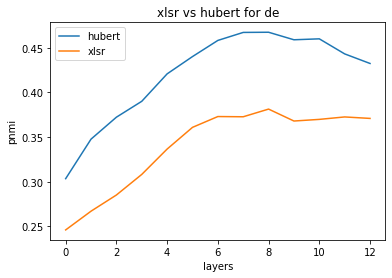}
        \caption{HuBERT vs. XLSR evaluated on German data}
    \end{subfigure}
    \hfill
    \begin{subfigure}{0.45\textwidth}
        \centering
        \includegraphics[width=\linewidth]{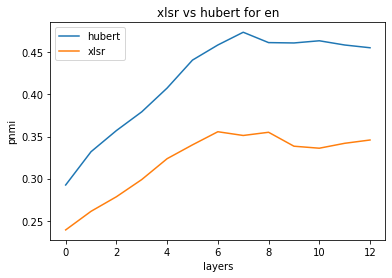}
        \caption{HuBERT vs. XLSR evaluated on English data}
    \end{subfigure}
    \begin{subfigure}{0.45\textwidth}
        \centering
        \includegraphics[width=\linewidth]{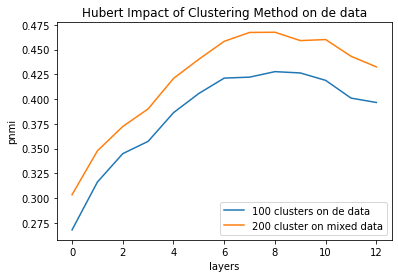}
        \caption{100 monolingual vs. 200 mixed units, evaluated on German data}
    \end{subfigure}
    \hfill
    \begin{subfigure}{0.45\textwidth}
        \centering
        \includegraphics[width=\linewidth]{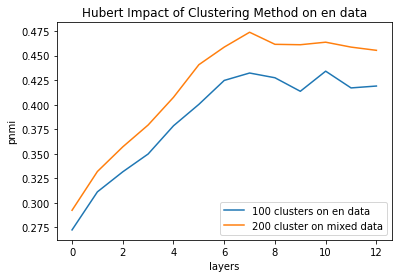}
        \caption{100 monolingual vs. 200 mixed units, evaluated on English data}
    \end{subfigure}
    \caption{PNMI vs. layer index, comparing different clustering settings for English and German. Higher is better.}
    \label{fig:pnmi}
\end{figure*}

\begin{figure}[ht!]
    \centering
    \includegraphics[width=\linewidth]{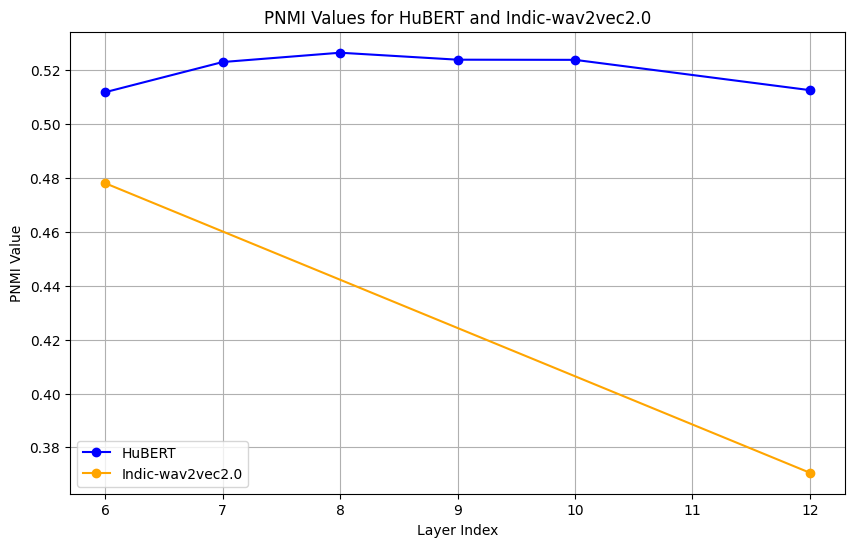}
    \caption{PNMI with HuBERT and Indic wav2vec2.0 evaluated on Shrutilipi, computed for different layer indices, for Marathi. Higher is better.}
    \label{fig:pnmimr}
\end{figure}

\section{S2U Encoder Ablations}\label{app:s2uresults}
We decide (a) which speech encoder model to use, (b) whether to learn separate per-language k-means models or a joint k-means model and (c) which encoder layer to take embeddings from, based on the average Pointwise Normalized Mutual Information (PNMI) between unit sequences and phoneme sequences extracted from the same datasets, following~\citet{Hsu2021HuBERTSS}. Our best configuration uses a single Marathi k-means model and a shared English-German k-means model. We find that this works better than training three individual models or a single model, which we hypothesize is due to language similarities.

To obtain the phoneme sequences for English and German, we use English and German phonemizers from the Montreal Forced Aligner\footnote{\url{https://montreal-forced-aligner.readthedocs.io/en/latest/}}. For Marathi, we use a Kaldi-based ASR model trained on Shrutilipi data. To train the k-means models, we use $\approx$ 50 hrs of speech data from each language, obtained from a random subset of Librispeech~\citep{librispeech} for English, MLS~\citep{Pratap_2020} for German, and Shrutilipi~\citep{bhogale2022effectiveness} for Marathi.

First, we describe our ablations for English-German. We experiment with different base speech models (HuBERT~\citep{Hsu2021HuBERTSS} vs. XLSR~\citep{conneau2020unsupervised}), layer indices, number of clusters ($100$ vs. $200$) and types of clusterings (one clustering for both languages jointly v.s. separate clusterings) and choose the configuration that achieves the highest Pointwise Normalized Mutual Information (PNMI).
We report PNMI results for these English-German configurations in Figure~\ref{fig:pnmi}.

For Marathi, we experiment with different base speech models (HuBERT vs Indic-wav2vec2.0~\citep{javed2021building}) and layer indices. We fix the number of clusters at $100$. We choose the configuration that achieves the highest PNMI. We report PNMI results for these Marathi configurations in Figure~\ref{fig:pnmimr}.

\section{U2S Modeling and Evaluation}
\label{app:resynthesis}
Using the unit sequences for the Voxpopuli (English and German) and Shrutilipi (Marathi) datasets, generated from our S2U encoder, we train vocoders to generate the speech from these unit sequences. We train across $4$ GPUs with a learning rate of $2e-4$ with a batch size of $128$ (for en-de) and $240$ (for mr) and train for $60$k updates; other hyperparameters follow~\citet{polyak21_interspeech}. As a sanity check, we evaluate S2U and U2S by computing the resynthesis WER, which measures how well passing a given speech signal through S2U and U2S preserves the content of the input speech signal.

We compute the resynthesis WER as follows: (1) pass input speech to the S2U encoder and generate the unit sequence, (2) pass the generated unit sequence to our U2S vocoder to synthesize speech, (3) transcribe the synthesized speech using ASR (4) compute the Word Error Rate between the transcript and the ground truth transcript of the input speech. To account for the errors from ASR, we compute the WER between the ASR transcript of the input speech utterance (`ground-truth' speech) and the ground truth transcript as a lower bound. We use test sets from English and German Voxpopuli~\citep{voxpopuli} and English LJSpeech~\citep{ljspeech17} with our synthetic single-speaker speech. Table~\ref{tab:vocoder} presents these results. We find that the resynthesis WERs are fairly good for English, and worse for German. Based on qualitative analysis of the German input speech (which is already single-speaker synthetic speech) and resynthesized speech (passed through S2U and U2S), we find that the input speech itself makes stress and pronunciation errors, driving up the Ground Truth WER, which further cascades into the model resynthesis WER. We still use this model because it is the best we could build with existing tools.
\begin{table}
\centering
\begin{tabular}{cccc}
\toprule
\textbf{Method} & \multicolumn{1}{c}{\textbf{en VP}} & \multicolumn{1}{c}{\textbf{de VP}} & \multicolumn{1}{c}{\textbf{en LJ}}\\
\midrule
Ground Truth & 4.89 & 8.44 & 3.80 \\
~\citep{lee-etal-2022-direct} & 10.56 & - & 7.69 \\
Ours & 8.53 & 19.46 & 6.72 \\
\bottomrule
\end{tabular}
\caption{S2U + U2S resynthesis performance; WER computed between resynthesized speech transcribed by ASR model and ground truth transcripts. Lower WER is better. We also include the ground-truth speech WER as a lower bound. VP = Voxpopuli, LJ = LJSpeech }\label{tab:vocoder}
\end{table}

\begin{table*}[t]
\centering
\setlength{\tabcolsep}{2pt}
\begin{tabular}{rlcccc}
\toprule
& & & \multicolumn{3}{c}{\textbf{ASR-BLEU} $\uparrow$} \\
\cmidrule(lr){4-6}
& & & \multicolumn{2}{c}{\textbf{Europarl-ST}} & \textbf{CVSS} \\
\cmidrule(lr){4-5} \cmidrule(lr){6-6}
& \textbf{Model} & \textbf{Parallel} \#\textbf{hrs} & \textbf{de$\rightarrow$en} & \textbf{en$\rightarrow$de} & \textbf{de$\rightarrow$en} \\
\midrule
\circled{o} & Cascaded ASR-MT~\citep{https://doi.org/10.48550/arxiv.1911.03167} & N/A & 21.3 & 22.4 & - \\
\circled{p} & E2E S2T~\citep{voxpopuli} & 226h & 17.5 & - & - \\
\circled{q} & E2E S2T w/ Voxpop-Aligned~\citep{voxpopuli} & $\approx$500h & 18.8 & - & - \\
\circled{r} & Translatotron 2~\citep{translatotron2} & 120h & - & - & 19.7 \\
\bottomrule
\end{tabular}
\caption{English-German translation evaluation using BLEU for topline S2T models (rows \circled{o}-\circled{q}) and ASR-BLEU for S2ST model, row \circled{r} on Europarl-ST~\citep{https://doi.org/10.48550/arxiv.1911.03167} test set; higher is better. The Parallel \#hrs column denotes the size of parallel translation training data.}
\label{tab:appendixresults}
\end{table*}

\section{Text-based, Parallel-High-Resource S2T/S2ST models}
For completeness, we describe existing text-based, parallel-high-resource models in the literature and showcase their results in Table~\ref{tab:appendixresults}. These models date to 2021 and underperform the text-based parallel-low-resource models in our main results (Table~\ref{tab:resultsfinal}) but outperform textless parallel-high-resource models. Rows \circled{o}-\circled{q} are S2T models while \circled{r} is an S2ST model. \circled{o}~\citep{https://doi.org/10.48550/arxiv.1911.03167} is an ASR-MT cascade model whose MT component is trained on a large-scale text translation dataset OPUS~\citep{tiedemann-2012-parallel}.  \circled{p} and \circled{q} are Transformer-based models from ~\citet{voxpopuli} trained on the union of Europarl-ST and CVSS (total duration 226h) with \circled{q} being additionally trained on $\approx$300h of Voxpopuli aligned speech translation data. \circled{r} is the Translatotron 2~\citep{translatotron2}, a spectrogram-to-spectrogram encoder-synthesizer model trained with text supervision for the decoder with 120h of German-English data and about 360h of aligned data in 3 other X-to-English language pairs.

\section{Example Outputs}
\label{app:examples}
We present example outputs from our models. First, we showcase 10 cherry-picked examples, 2 examples from each evaluated language pair and domain in Table~\ref{tab:cherry}. Our best models, the post-backtranslation models (rows \circled{g} and \circled{n} in Tables~\ref{tab:resultsfinal} and~\ref{tab:resultsfinalmarathi}) perform well on these examples. We present the ground-truth transcripts of the source and target utterances, the ASR transcript of the target utterance predicted by the pre-backtranslation finetuned models (rows \circled{f} and \circled{m} in Tables~\ref{tab:resultsfinal} and~\ref{tab:resultsfinalmarathi}) and the ASR transcript of the target utterance predicted by our best models, the post-backtranslation models. We can observe that our post-backtranslation models are able to nearly perfectly translate these cherry-picked examples, which can be categorized into examples with (a) no mistakes (rows 1, 5, 7, 9), (b) valid replacements that largely preserve sentence meaning (rows 2, 4, 8) and (c) minor pronunciation errors (rows 6, 10). On the other hand, predictions from the finetuned model are overall worse, categorized into (a) no mistakes (row 1), (b) valid meaning-preserving replacements (row 2), (c) large meaning changes (row 3, 4, 7, 9, 10) and (d) incoherent output (row 5, 6, 8).

We also sample 5 randomly-picked examples, one from each setting to again compare our pre-backtranslation finetuned models and our best post-backtranslation models in Table~\ref{tab:random}. The examples show that the models are getting several of the words and semantics right, but often mistranslate certain words and make egregious grammatical and language modelling mistakes. We can see that our post-backtranslation model is overall better than the finetuned model for English-German in row (1), (2), worse in row (3), and performs similarly for rows (4) and (5).

\begin{table*}[!htp]\centering
\scriptsize
\setlength{\tabcolsep}{0.5em}
\begin{tabular}
{rp{0.2\linewidth}p{0.2\linewidth}p{0.2\linewidth}p{0.2\linewidth}}\toprule
 & \textbf{Source Utterance} & \textbf{Target Utterance (Gold)} & \textbf{Prediction from finetuned model} & \textbf{Prediction from post-backtranslation model} \\\midrule
 & \textbf{\textit{en$\rightarrow$de (Europarl-ST)}} & & & \\
 (1) & you can take initiatives & sie können initiativen ergreifen & sie können initiativen ergreifen & sie können initiativen ergreifen \\
\cdashlinelr{1-5}
(2) & madam president i \underline{supported} this report & frau präsidentin ich habe diesen bericht \underline{unterstützt} & frau präsidentin ich unterstütze diesen bericht & frau präsidentin ich habe diesen bericht \underline{gestimmt} \\
\midrule
 & \textbf{\textit{de$\rightarrow$en (Europarl-ST)}} & & & \\
(3) & ich denke da sind wir auf dem richtigen weg & i think we are on the right track \underline{here} & i think we should be aware of this & i think we are on the right track\\
\cdashlinelr{1-5}
(4) & ich denke es ist klar dass die bürger und bürgerinnen der europäischen union \underline{diese steuer} wollen und ich denke dass \underline{es eine große} \underline{verantwortung ist} & i think it is clear that the citizens of the european union want \underline{this tax} and i think \underline{we have a great} \underline{responsibility here} & i think that it is clear that the citizens of the european union want to do with these tasks and to do with the european union what it wants to do & i think it is clear that the citizens of the european union want \underline{to be taxed} and i think \underline{it is a major responsibility} \\
\midrule
 & \textbf{\textit{de$\rightarrow$en (CVSS)}} & & & \\
(5) & stellst du die musik bitte auf zimmerlautstärke albert rief seine mutter & are you turning the volume down to room volume albert his mother screamed & are you turning the music albert towards its mountain rock & are you turning the volume down to room volume albert his mother screamed \\
\cdashlinelr{1-5}
(6) & \underline{los} angeles liegt an der westküste & \underline{los} angeles is located on the west coast & loosen hot air line at the west coast & \underline{rose} angeles is located on the west coast \\
\midrule
 & \textbf{\textit{mr$\rightarrow$en (S-EP-ST)}} & & \\
(7) & {\dn yA kArZA\2\7{m}\30Fw\? mF yA ahvAlAQyA bA\8{j}n\? mt d\?U fkt nAhF} & for these reasons i cannot vote in favour of this report & for this reason i am in favour of the report & for these reasons i cannot vote in favour of this report \\
\cdashlinelr{1-5}
(8) & {\dn t\? aADFc \underline{\7{s}DAErt k\?l\? g\?l\? aAh\?} pr\2\7{t} aAZKF kAm krZ\? aAv\35Bwyk aAh\?} & it has already \underline{been modified} but more work needs to be done & it is improving barrowness improving but it must be forgotten & it has already \underline{made improvements} but more work needs to be done \\
\midrule
 & \textbf{\textit{mr$\rightarrow$en (S-Shr-ST)}} & & \\
(9) & {\dn p\2c\?cA\30FwFs vqA\0vrQyA svA{\rdt}nF lsFkrZ av\35Bwy kzn \35DwyA} & all those above forty five years must get vaccinated & more than forty five years of vaccination papers & all those above forty five years must get vaccinated \\
\cdashlinelr{1-5}
(10) & {\dn t\? kAl \underline{\7{m}\2b\4it} bAtmFdArA\2fF bolt hot\?} & he was talking to reporters in \underline{mumbai} yesterday & he was talking to reporters in mabay to day & he was talking to reporters in \underline{mumba} yesterday \\
\bottomrule
\end{tabular}
\caption{Cherry-picked examples picked for our best S2ST models (the post-backtranslation models), reporting predictions for both finetuned and post-backtranslation models. We manually annotate the differences between the gold utterance and the prediction from the post-backtranslation model, align them to the source utterance and underline the differences.} \label{tab:cherry}
\end{table*}

\begin{table*}[!htp]\centering
\scriptsize
\begin{tabular}
{rp{0.2\linewidth}p{0.2\linewidth}p{0.2\linewidth}p{0.2\linewidth}}\toprule
 & \textbf{Source Utterance} & \textbf{Target Utterance (Gold)} & \textbf{Prediction from finetuned model} & \textbf{Prediction from post-backtranslation model}\\
 \midrule
 & \textbf{\textit{en$\rightarrow$de (Europarl-ST)}} & & & \\
 (1) & goods and cargo have been delayed or not transported at all and businesses both large and small have been affected & waren und güterlieferungen wurden verschoben oder ganz gestoppt und sowohl kleine als auch große unternehmen sind betroffen & kosovo und konsum wurden zerstört oder wurden nicht erwähnt oder angemessen sein können & günstige und kunden wurden im vorle von kmos nicht erwähnt oder noch nicht erwähnt von allen unternehmen großen unternehmen\\
\midrule
 & \textbf{\textit{de$\rightarrow$en (Europarl-ST)}} & & & \\
(2) & wir sollten hier nicht mit zweierlei maß messen & we must not apply double standards here & we should not do so with these matters & we should not be here with the two sides\\
\midrule
 & \textbf{\textit{de$\rightarrow$en (CVSS)}} & & &\\
(3) & ihr schalldeckel trägt herabhängende quasten und ist mit einem pelikan bekrönt & their sounding board has loose hanging tassels and is crowned with a pelican & year study teacher however remaining costs and an ice and hobbies & child dictatorial territorial castes and is managed by a pellikov\\
\midrule
 & \textbf{\textit{mr$\rightarrow$en (S-EP-ST)}} & & & \\
(4) & {\dn n\4sEg\0k s\2sADn\? aAEZ EnsgA\0c\? s\2r\322wZ kr\317wyAsAWF aApSyAlA pyA\0vrZ s\2r\322wZAQyA \322w\?/At s\2vAdAcF aAv\35BwyktA aAh\?} & we need dialogue in the field of environmental protection in order to conserve natural resources and nature & in order to protect natural resources and defense quality basis we need a clear signal of environmental protection & we need collectively in the area of protection resources for natural resources and jobs\\
\midrule
 & \textbf{\textit{mr$\rightarrow$en (S-Shr-ST)}} & & & \\
(5) & {\dn \7{m}\2b\4i aAEZ upngrA\2m@y\? g\?SyA kAhF EdvsA\2t jordAr pAUs JASyA\7{m}\30Fw\2 sAt \7{m}Hy tlAvA\2QyA pA\317wyAt l\322wZFy vAY JASyAn\2 \7{m}\2b\4ilA \7{p}YFl bArA mEhn\? pAZF \7{p}rvWA \7{s}r\30FwFtpZ\? hoU fkZAr aAh\?} & heavy rains in mumbai and its suburbs in the last few days have significantly increased the water level in the seven main lakes ensuring smooth water supply to mumbai for the next twelve months & in the last few days ero people who have done in mumba mumbai soon reins have done in the last few days in the last few days mumbai & in mumba and opportunities of mumba and mumba who have received water in seventeen t h needs water in the last few days by the water in the mumbai\\
\bottomrule
\end{tabular}
\caption{Randomly sampled examples comparing our finetuned and post-backtranslation models.}\label{tab:random}
\end{table*}

\end{document}